\documentclass[lettersize,journal]{IEEEtran}
\usepackage{graphicx}
\usepackage{amsmath}    % 数学公式
\usepackage{amssymb}    % 更多数学符号
\usepackage[ruled,vlined,linesnumbered]{algorithm2e}
\SetCommentSty{mycommfont}
\SetKwComment{CommentLine}{\color{blue}\small /$\ast$ }{ $\ast$/}
\usepackage{xcolor}
\DontPrintSemicolon  % 去掉每行末尾的分号

\usepackage{threeparttable}

\usepackage{mathtools}
\usepackage{mathrsfs}
\usepackage{pifont}

\usepackage{gensymb,soul}
\usepackage{graphics} % for pdf, bitmapped graphics files
\usepackage{subfigure}
\usepackage{booktabs}
\usepackage{epsfig} % for postscript graphics files
\usepackage{cite}
\usepackage{mathptmx} 
\usepackage{amsfonts,amssymb, amsmath}  
\usepackage{float}  % 引入 float 宏包
\usepackage[colorlinks,linkcolor=blue]{hyperref}

\usepackage{xcolor}
\usepackage{ifthen}
\usepackage{makecell}
\usepackage{algorithmicx}
\usepackage{algpseudocode}
\usepackage{threeparttable}
\usepackage[linesnumbered,ruled]{algorithm2e}
\usepackage{booktabs}

\DeclareMathAlphabet{\mathcal}{OMS}{cmsy}{m}{n}
\DeclareSymbolFont{largesymbols}{OMX}{cmex}{m}{n}
\usepackage{multirow}

\begin{document}

\title{\LARGE \bf
ALORE: Autonomous Large-Object Rearrangement \\ with a Legged Manipulator
}
% whole-body controller (WBC)

% \author{Author Names Omitted for Anonymous Review}
% \author{Zhihai Bi, Yushan Zhang, Kai Chen, Guoyang Zhao, Yulin Li, and Jun Ma, \textit{Senior Member, IEEE} % stops a space

%  \thanks{Zhihai Bi, Yushan Zhang, Kai Chen, Guo are with the Robotics and Autonomous Systems Thrust, The Hong Kong University of Science and Technology (Guangzhou), Guangzhou 511453, China (e-mail: zbi217@connect.hkust-gz.edu.cn). }
% \thanks{Yulin Li and Jun Ma are with the Robotics and Autonomous Systems Thrust, The Hong Kong University of Science and Technology (Guangzhou), Guangzhou 511453, China, and also with the Division of Emerging Interdisciplinary Areas, The Hong Kong University of Science and Technology, Hong Kong SAR, China (e-mail: jun.ma@ust.hk). \textit{(Corresponding author: Jun Ma).}}
% }%

\author{Zhihai Bi, Yushan Zhang, Kai Chen, Guoyang Zhao, Yulin Li, and Jun Ma, \textit{Senior Member, IEEE} % stops a space

\thanks{Zhihai Bi, Yushan Zhang, Kai Chen, Guoyang Zhao, Yulin Li, and Jun Ma are with the Robotics and Autonomous Systems Thrust, The Hong Kong University of Science and Technology (Guangzhou), Guangzhou 511453, China.}
}

% \markboth{IEEE Transactions on Robotics}%


\makeatletter
\newcommand{\TitleFig}[2]{% #1=caption #2=label
  \begingroup
  \def\@captype{figure}\caption{#1}\label{#2}%
  \endgroup
}
\makeatother

\IEEEaftertitletext{%
\vspace{-3.0em}
\begin{center}
\includegraphics[width=\linewidth]{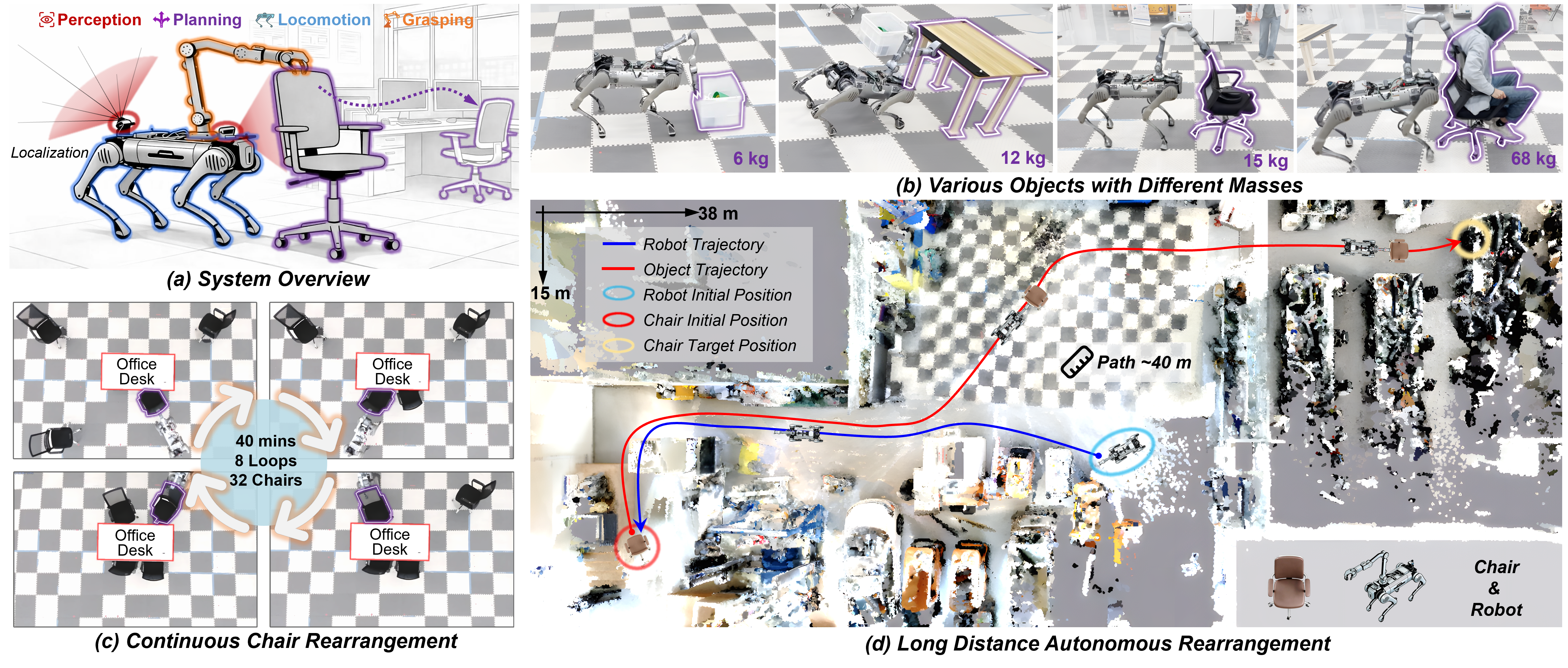}
\TitleFig{%
The proposed system, ALORE, seamlessly integrates perception, planning, locomotion, and grasping, enabling a legged manipulator to perform diverse large-object rearrangement tasks. (b) It achieves stable grasping and accurate planar velocity control across various objects with different masses, while preventing self-collision and object toppling. (c) The system also demonstrates long-term autonomy, running continuously for 40 minutes to complete 8 loops and totally rearrange 32 chairs without a single failure. (d) Moreover, ALORE supports long-distance autonomous rearrangement, moving a chair along an approximately 40\,m route through narrow passages without collisions, over uneven terrain with protrusions, and across floor surfaces of varying materials.%
}{fig:cover}
\end{center}
\vspace{-2pt}%
}

% \IEEEaftertitletext{%
% \vspace{-3.0em} % 
% \noindent\begin{center}
% \includegraphics[width=\linewidth]{images/cover-fig.pdf}
% \captionof{figure}{
% The proposed system, ALORE, seamlessly integrates perception, planning, locomotion, and grasping, enabling a legged manipulator to perform diverse large-object rearrangement tasks. (b) It achieves stable grasping and accurate planar velocity control across various objects with different masses, while preventing self-collision and object toppling. (c) The system also demonstrates long-term autonomy, running continuously for 40 minutes to complete 8 loops and totally rearrange 32 chairs without a single failure. (d) Moreover, the system supports long-distance autonomous rearrangement, successfully moving a chair along an approximately 40\,m route through narrow passages without collisions, over uneven terrain with protrusions, and across floor surfaces of varying materials.
% }
% \label{cover-figure}
% \end{center}
% % \vspace{0.5em} %
% \vspace{-5pt}
% }

\maketitle

\begin{abstract}
Endowing robots with the ability to rearrange various large and heavy objects, such as furniture, can substantially alleviate human workload.
However, this task is extremely challenging due to the need to interact with diverse objects and efficiently rearrange multiple objects in complex environments while ensuring collision-free loco-manipulation.
In this work, we present ALORE, an \underline{A}utonomous \underline{L}arge-\underline{O}bject \underline{RE}arrangement system for a legged manipulator that can rearrange various large objects across diverse scenarios.
The proposed system is characterized by three main features: (i) a hierarchical reinforcement learning training pipeline for multi-object environment learning, where a high-level object velocity controller is trained on top of a low-level whole-body controller to achieve efficient and stable joint learning across multiple objects; (ii) two key modules, a unified interaction configuration representation and an object velocity estimator, that allow a single policy to regulate planar velocity of diverse objects accurately; and (iii) a task-and-motion planning framework that jointly optimizes object visitation order and object-to-target assignment, improving task efficiency while enabling online replanning.
Comparisons against strong baselines show consistent superiority in policy generalization, object-velocity tracking accuracy, and multi-object rearrangement efficiency.
Key modules are systematically evaluated, and extensive simulations and real-world experiments are conducted to validate the robustness and effectiveness of the entire system, which successfully completes 8 continuous loops to rearrange 32 chairs over nearly 40 minutes without a single failure, and executes long-distance autonomous rearrangement over an approximately 40\,m route.
The open-source packages are available at \url{https://zhihaibi.github.io/Alore/}.
\end{abstract}

\begin{IEEEkeywords}
Loco-manipulation, whole-body control, reinforcement learning, object rearrangement.
\end{IEEEkeywords}

%%%%%%%%%%%%%%%%%%%%%%%%%%%%%%%%%%%%%%%%%%%%%%%%%%%%%%%%%%%%%%%%%%%%%%%%%%%%%%%%
\section{INTRODUCTION}
% \IEEEPARstart{O}{bject} rearrangement objects are a fundamental and indispensable capability in daily life. Endowing robots with this competency can significantly alleviate human workload, particularly in scenarios involving large and heavy objects \cite{gu2025humanoid, barreiros2025learning}. For instance, in domestic environments, robots can rearrange furniture such as chairs and tables \cite{li2025robotmover}, while in industrial settings, they can undertake logistics-related tasks such as goods sorting and organization \cite{murooka2021humanoid, chappellet2023humanoid}.

\IEEEPARstart{O}{bject} manipulation is a cornerstone of robotics, enabling robots to perform meaningful work in complex, unstructured human environments.\cite{billard2019trends, ai2025review}. Despite decades of progress, most advances have primarily been made in tabletop manipulation of small and lightweight objects in structured environments \cite{batra2020rearrangement, xu2024grasp, andrychowicz2020learning}. Yet many practical applications require manipulation at a much larger physical scale, where robots must interact with bulky objects that are infeasible to lift directly. A representative example is large-object rearrangement \cite{ravan2024combining}, in which objects must be rearranged to desired configurations through contact-rich interactions, such as pulling or pushing, as commonly seen in office settings when organizing chairs and tables. Moreover, the challenge becomes even more pronounced in complex scenes such as warehouses and logistics facilities, where diverse objects must be rearranged over long distances and long time horizons. Endowing mobile robots, particularly legged manipulators, with such a scalable large-object rearrangement capability can substantially reduce human physical workload and, more importantly, push the limits of robotic loco-manipulation in real-world scenarios.

%  \begin{figure}[t]	
% 	\centering
%     % \vspace{-10pt}
% 	\includegraphics[width=0.99\linewidth]{images/head2.jpg}
% 	% \setlength{\abovecaptionskip}{-2pt} 
% 	\caption
% 	{The proposed system enables a legged manipulator to perform diverse large-object rearrangement tasks. (a) Planar velocity control across object categories under varying payloads (b) Top-down view of the multi-chair experiment. The system operated continuously for nearly 40 minutes, successfully completing eight loops and rearranging a total of 32 chairs.}
%     \label{fig:head}
%     \vspace{-12pt}
% \end{figure}

In recent years, several approaches have been proposed for loco-manipulation tasks involving interactions with large objects. They can be broadly categorized into model-based approaches and learning-based approaches \cite{gu2025humanoid}. Model-based methods typically rely on explicitly modeling the dynamics of the coupled robot-object system and leveraging this model within a model predictive control framework to generate adaptive loco-manipulation behaviors \cite{audren2014model}. However, whether modeling the object as an external wrench \cite{murooka2015whole} or adopting a unified robot-object model \cite{8528498}, these approaches require perfect object knowledge (e.g., mass and friction coefficient), which is usually unavailable in practice, especially in rearrangement tasks involving diverse objects.

In contrast, model-free methods, particularly reinforcement learning (RL), have emerged as a promising alternative \cite{ha2025learning}. By leveraging sampling interaction data in a high-fidelity simulator, RL-based approaches learn neural control policies, i.e., state-to-action mappings, without requiring explicit derivation of system dynamics \cite{he2025attention}.
For large-object rearrangement, existing RL-based approaches can be grouped into two paradigms. 
One line of work trains end-to-end policies that map perception and robot-object states directly to joint-level commands. However, learning such policies that simultaneously generalize across manipulation and navigation remains challenging because the long-horizon credit assignment under sparse rewards \cite{luo2025precise}, and most prior methods are restricted to a single object and a single scene \cite{mendonca2024continuously}.
Another paradigm adopts an object-centric formulation, such as learning an object velocity controller, which enables the object to follow a planned, collision-free trajectory  \cite{ravan2024combining}. This hybrid paradigm offers greater flexibility and has demonstrated improved transferability across different environments \cite{li2025robotmover}.
Nevertheless, when extended to multi-object rearrangement, these hybrid approaches still exhibit several limitations. First, for the RL training pipeline, existing methods can only collect interaction data in single-object environments \cite{bi2025interactive}, preventing joint training across multiple object categories in a unified setup. Second, at the policy level, previous studies mainly learn policies that are only applicable to a single object, which limits generalization across diverse objects \cite{ravan2024combining}. Third, from a system-level perspective, existing approaches typically consider only single-object rearrangement and still rely on teleoperation \cite{li2025robotmover} or external sensing \cite{mendonca2024continuously}, and a complete rearrangement system for various large objects has yet to be demonstrated.

\begin{table}[t]
    \centering
    \caption{Comparison with Related Works}
    \label{tab:related_work}
    \begin{tabular}{ccccc}
        \toprule
        \textbf{Method} &
        \makecell{Velocity \\ Control} &
        \makecell{Various \\ Objects} &
        \makecell{Various \\ Scenarios} &
        \makecell{Complete \\ System} \\
        \midrule
        Yang et al. \cite{yang2025efficient}  & \textcolor{blue}{\ding{55}}  & \textcolor{blue}{\ding{55}} & \textcolor{green}{\ding{51}} & \textcolor{blue}{\ding{55}} \\
        
       Dadiotis et al. \cite{dadiotis2025dynamic} & \textcolor{blue}{\ding{55}}  & \textcolor{green}{\ding{51}} & \textcolor{blue}{\ding{55}} & \textcolor{blue}{\ding{55}} \\
       
        Ravan et al. \cite{ravan2024combining} & \textcolor{blue}{\ding{55}}  & \textcolor{blue}{\ding{55}} & \textcolor{blue}{\ding{55}} & \textcolor{blue}{\ding{55}}\\
        
        Li et al. \cite{li2025robotmover} & \textcolor{green}{\ding{51}} & \textcolor{blue}{\ding{55}} & \textcolor{green}{\ding{51}} & \textcolor{blue}{\ding{55}}\\
        
        \textbf{Our Method} & \textcolor{green}{\ding{51}}  & \textcolor{green}{\ding{51}} & \textcolor{green}{\ding{51}} & \textcolor{green}{\ding{51}}\\
        \bottomrule
    \end{tabular}
    \vspace{-10pt}
\end{table}

To overcome the aforementioned limitations, we develop a complete large-object rearrangement system for a legged manipulator based on a hybrid paradigm. Specifically, we build a multi-object simulation environment that enables the simultaneous collection of interaction data across diverse objects. Leveraging this environment, we develop a hierarchical RL pipeline for multi-object learning, where a high-level object velocity controller is trained above a low-level whole-body controller (WBC) to generate action sequences under safety constraints.
Moreover, inspired by physical interaction modeling in the animation domain \cite{zhang2023simulation}, we propose a unified and generalizable representation for robot-object interaction, termed the interaction configuration representation (ICR). Specifically, we leverage graph structure to model diverse interaction configurations between the robot and objects, enabling a single policy to more effectively learn and generalize across interactions with different kinds of objects. 
To further improve object velocity tracking accuracy in a closed-loop manner, we learn an object velocity estimator that can accurately predict various objects' planar velocity.
Additionally, we propose a TAMP framework for efficient multi-object rearrangement that jointly optimizes object visitation order and target pose assignment to minimize completion time, and supports online replanning via coarse-to-fine trajectory generation.
By integrating multi-object rearrangement planning, tailored perception modules, and hierarchical RL-based control, the proposed system enables long-horizon rearrangement of diverse large objects across varied environments. Comparisons with related approaches are presented in Table~\ref{tab:related_work}.

Overall, the main contributions are summarized as follows:
\begin{enumerate}

\item A complete large-object rearrangement system for a legged manipulator is introduced. To the best of the authors’ knowledge, this is the first system to achieve long-horizon, long-distance rearrangement of diverse large objects across a wide range of scenarios.

\item A hierarchical RL training pipeline is proposed to learn generalizable interactions across various objects. The high-level object controller, augmented with an ICR module and an object velocity estimator, is trained on top of a low-level WBC to achieve stable object interaction with safety constraints, enabling a single policy to accurately regulate planar velocities across diverse objects.

% \item A hierarchical RL training pipeline is proposed for multi-object environments learning, where a high-level object velocity controller is trained on top of a low-level WBC to achieve stable interaction with multiple objects under safety constraints.

% \item Both the ICR and the object velocity estimator are designed to encode interaction configuration and predict object velocities, respectively, allowing a single policy to accurately regulate planar velocity of different objects.

\item A TAMP framework is developed for efficient multi-object rearrangement, which optimizes the overall task completion time and supports online replanning via coarse-to-fine trajectory generation.

\item Extensive experiments are conducted to validate the entire system in large-scale scenarios, highlighted by a long-horizon trial that continuously rearranges 32 chairs over nearly 40 minutes and a long-distance autonomous rearrangement task spanning nearly 40\,m.
\item All system components, including the RL-based training pipeline in simulation and the real-world deployment framework, are released as open-source packages to facilitate future development.
\end{enumerate}

The rest of the article is organized as follows. In Sec. \ref{sec:related work}, we review the related literature about large-object rearrangement. In Sec. \ref{sec:overview}, the overview of the system architecture is given. Then, we introduce the training details of the RL-based hierarchical controller in Sec. \ref{sec:hie}, including the low-level WBC and the high-level object controller. Next, the tailored TAMP algorithm for multi-object rearrangement is demonstrated in Sec. \ref{sec:TAMP}. In Sec. \ref{sec:exp}, we present extensive results from both simulation and real-world experiments and provide a comprehensive analysis. Limitations are discussed in Sec. \ref{sec:limit}, and conclusions are drawn in Sec. \ref{sec:conclusion}.

\section{Related Work}\label{sec:related work}

\subsection{Large Object Rearrangement}
Although significant progress has been made in tabletop rearrangement for lightweight, small-sized objects  \cite{batra2020rearrangement, xu2024grasp, zhao2024fots, suomalainen2022survey}, research on manipulating objects with a large size remains limited. Such objects are typically too heavy or bulky to be lifted by a single arm and instead must be moved via contact-rich actions such as pushing or pulling along the ground, making it substantially more challenging.

In recent years, several works have started to tackle large-object rearrangement, typically using either end-to-end learning or hybrid pipelines that combine learning-based components with classical planning to rearrange objects such as chairs \cite{mendonca2024continuously, dai, wang2024interactive}, tables \cite{li2025robotmover, xia2020interactive}, and boxes \cite{ yang2025efficient, ellis2022navigation, figueroa2020dynamical}. The most relevant prior works include \cite{mendonca2024continuously, ravan2024combining, li2025robotmover}, which employ legged manipulators to accomplish long-horizon large-object rearrangement tasks.
Mendonca et al. \cite{mendonca2024continuously} present an end-to-end real-world RL framework that uses an RRT*-planned path as a prior and injects it into the replay buffer to bootstrap learning. It not only accelerates convergence but also enables adaptive interaction behaviors, such as recovering from collisions. However, the policy’s inputs (e.g., object semantics) rely on an external global camera. 
Unlike \cite{mendonca2024continuously}, PoPi \cite{ravan2024combining} follows a hybrid pipeline. It first collects teleoperated demonstrations to train a local object controller that maps the chair state and perceptual observations to robot actions. Then, the local controller is utilized to track waypoints generated by an A* algorithm. Since the demonstrations are collected in a single scene, the learned policy is difficult to generalize to different environments. 
Further, RobotMover \cite{li2025robotmover} leverages human-object interaction demonstrations to construct a reward function and trains an object velocity controller to track planned object paths, demonstrating obstacle avoidance across multiple scenarios.
In general, the above methods have enabled long-distance rearrangement of large objects. However, from a system-level perspective, most of them still focus on single-object rearrangement and pay limited attention to multi-object tasks. Moreover, existing systems are often incomplete: some studies still rely on teleoperation for object grasping \cite{li2025robotmover}, while others rely on external sensors \cite{mendonca2024continuously}.

In this work, we adopt a hybrid paradigm and develop a complete large-object rearrangement system that seamlessly integrates tailored perception, multi-object TAMP, and an RL-based controller to enable long-horizon, long-distance rearrangement across diverse large objects and scenarios.

\subsection{Loco-Manipulation with Large Objects}
For legged robots equipped with manipulators, loco-manipulation of large objects is highly valuable for assisting humans, and is a compelling research direction \cite{agravante2019human}.
In this task, the robot must coordinate locomotion and manipulation to control the object’s motion as desired.
Over the past decade, substantial efforts have been devoted to developing model-based approaches for achieving loco-manipulation skills \cite{ferrari2023multi, 10156397, rouxel2022multicontact, adu2023exploring, murooka2021humanoid, henze2016passivity}.
Some works treat the object’s contact effect as a known external wrench and incorporate it into the MPC to compute control inputs \cite{polverini2020multi}. This formulation is simple and easy to integrate, but the external wrench must be predefined. While other works build a unified robot-object model by augmenting the dynamics with object states \cite{audren2014model}, they still require reliable object physical information, including mass, friction coefficient, and so on. As a result, their deployment in open-world settings with novel objects of unknown physical properties remains challenging, substantially constraining their real-world applicability.

In recent years, learning-based methods have made substantial progress in loco-manipulation, covering tasks such as carrying heavy boxes \cite{yang2025omniretarget, yin2025visualmimic, zheng2025embracing}, opening doors \cite{xue2025opening, weng2025hdmi, chen2025chip}. 
OmniRetarget \cite{yang2025omniretarget} leverages the interaction mesh \cite{ho2010spatial} to generate high-quality training data, enabling proprioceptive RL policies to execute impressive platform-climbing skills. HDMI \cite{weng2025hdmi} synthesizes robot-object interaction data from monocular RGB videos and then applies RL to co-track the resulting trajectories to acquire skills such as box pushing. These methods either perform pure trajectory tracking with little tolerance for deviation from a predefined path or are restricted to short-horizon manipulation, and thus are not well suited for long-horizon (over 10 minutes) rearrangement tasks.
Mendonca et al. \cite{mendonca2024continuously}, and PoPi \cite{ravan2024combining} both use a robust RL-based controller for low-level locomotion, and then collect training data by interacting stably with objects using the manipulator. Similarly, Dadiotis et al. \cite{dadiotis2025dynamic} and RobotMover \cite{li2025robotmover} adopt a hierarchical RL training pipeline in simulation: they first pre-train a locomotion controller, and then train a goal-oriented pushing policy and an object-velocity controller, respectively. This hierarchical design ensures stable locomotion, allowing the learning to focus more on robot-object interaction.
Nevertheless, from a training pipeline perspective, most of them only collect interaction data within the same environment. As a result, the policies remain specialized to a single object, hindering generalization across diverse objects.

In this work, we construct a multi-object RL training environment that enables simultaneous collection of interaction data with diverse objects. Building on this environment, we adopt a hierarchical RL training pipeline and learn a high-level object velocity controller on top of a low-level WBC to achieve efficient and stable interaction with various objects.

 \begin{figure*}[t]	
	\centering
        % \vspace{10pt}
	\includegraphics[width=0.98\linewidth]{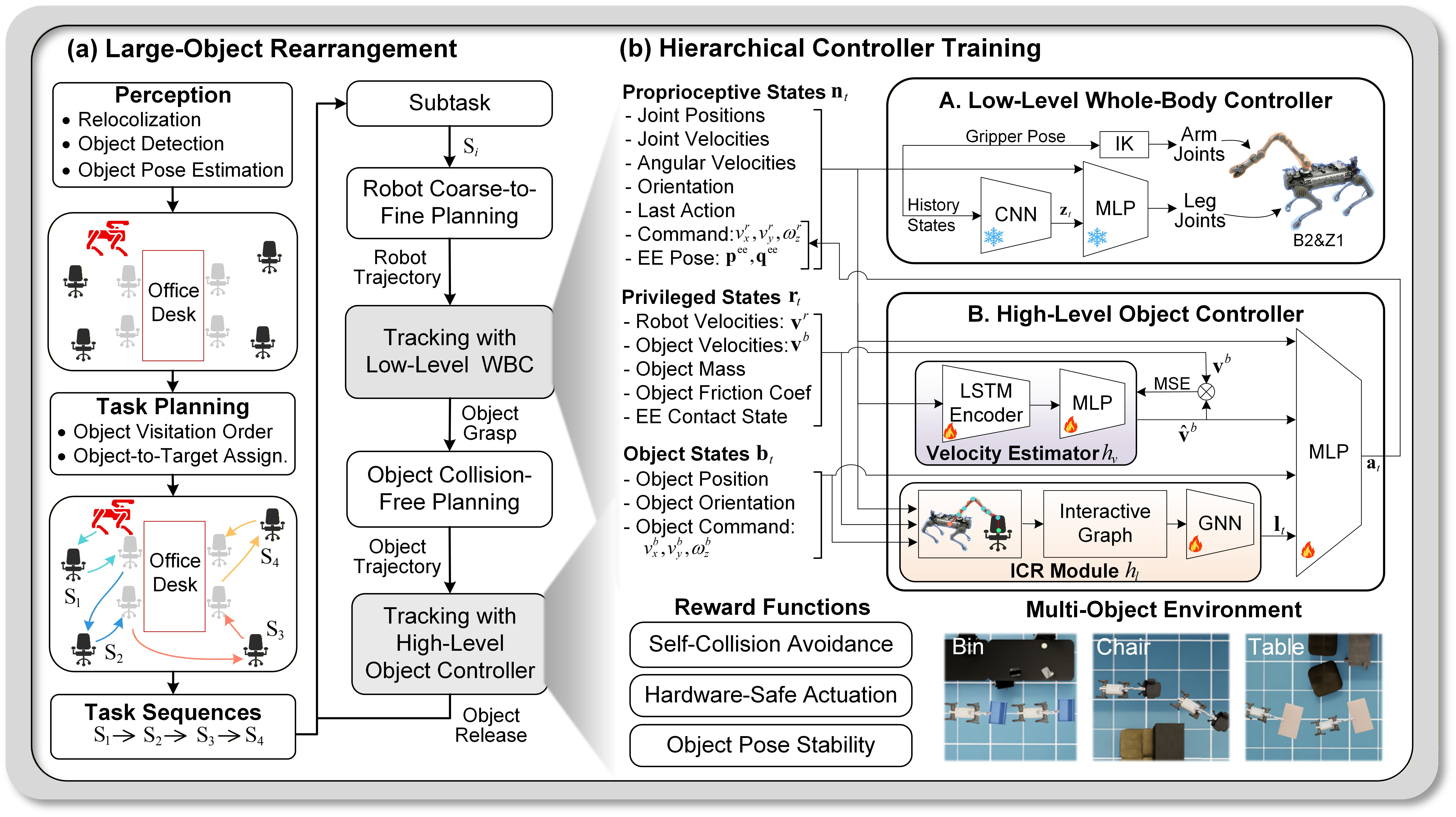}
	\setlength{\abovecaptionskip}{-2pt} 
	\caption
	{
    Illustration of a large-object rearrangement system with a legged manipulator. (a) Workflow of the proposed system, including tailored perception, robot-object system planning, and a hierarchical RL-based controller. (b) Training pipeline of the hierarchical controller, which consists of the low-level WBC and the high-level object controller. After training, the learned policy can be directly deployed on a legged manipulator without further fine-tuning.}
    \label{fig:1}
    \vspace{-8pt}

\end{figure*}

\subsection{Graph Representations for Physical Interaction}
% Robot-object interaction is inherently relational, shaped not only by the states of individual entities but also by their spatial couplings, contact connectivity, and functional constraints \cite{11127955}.
Graph representations provide a powerful way to model physical interaction configurations by explicitly encoding different entities as nodes and their relations (e.g., kinematic coupling) as edges with attributes \cite{soleymanzadeh2025simpnet}.

Recently, graph-based formulations have been adopted to describe robot-object interactions in a variety of settings \cite{sanchez2018graph, pfaff2020learning, allen2023graph}. 
Zhang et al.~\cite{zhang2023simulation} represent the interactions among body parts of articulated characters using an interaction graph (IG) to capture their spatial relationships.
RoboBallet \cite{lai2025roboballet} exploits the graphs to represent multirobot and objects, then uses graph neural networks (GNNs) for state-action value estimation.
Similarly, SIMPNet \cite{soleymanzadeh2025simpnet} leverages graphs and GNNs to retain the spatial relationships within the configuration space and the kinematic chain of the manipulator, leading to efficient motion planning.
We observe that graph-structured representations are particularly effective for capturing physical interactions, especially when the interaction exhibits a clear kinematic-chain structure \cite{ai2025review}. Meanwhile, we find that GNNs provide an effective way to encode such graphs via message passing and aggregation, explicitly incorporating entities, relations, and interaction rules into both the representation and computation, thereby enabling structured reasoning. \cite{battaglia2018relational}.

% Motivated by the observation above, we propose an ICR module that models the interaction configuration between the legged manipulator and the object as a graph. We further employ GNNs to extract graph-level features, which are provided as inputs to the high-level policy, enabling it to generate object-adaptive joint-level commands. The insight is that different robot-object interaction configurations require different joint-level commands for stable manipulation. We therefore encode the configuration and condition the high-level policy on this encoding to produce object-adaptive action distributions.

Motivated by the observation above, we propose an ICR module that models robot-object interaction configurations as a graph, providing a unified representation of robot-object interaction. Leveraging this representation, we employ GNNs to extract graph-level features and condition the high-level object policy to produce configuration-adaptive actions for stable object control.

\section{Overview of System Architecture}\label{sec:overview}

In this work, our goal is to autonomously rearrange various kinds of large objects, such as chairs, tables, and bins, to their target positions and orientations using a legged manipulator. To this end, we develop a complete large-object rearrangement system, as described in Fig. \ref{fig:1}(a). It seamlessly integrates tailored perception, multi-object TAMP, and an RL-based object controller. First, given objects' target poses, the task planning algorithm generates task sequences ($\text{S}_1 \rightarrow \text{S}_2 \rightarrow \cdot \cdot \cdot \rightarrow \text{S}_i$), including the object visitation order and the object-to-target assignment. Next, the coarse-to-fine robot trajectories for each subtask $\text{S}_i$ are computed based on the estimated object pose, which are then tracked by the robot's low-level WBC. Once the robot reaches the target pose, it grasps the object, and a collision-free trajectory is planned for the object. This trajectory is tracked by the high-level object controller until the object is released at its goal pose. The whole process above is repeated until all subtasks are finished. In this task, the perception module provides global poses of the robot and objects. Similar to prior large-object rearrangement settings \cite{li2025robotmover}, we assume that the approximate positions of target objects are available (without requiring precise poses) and that feasible grasp regions are provided. It allows us to focus on the core challenges, i.e., robust post-grasp interaction control across diverse large objects.

\section{Training Pipeline of Hierarchical Reinforcement Learning}\label{sec:hie}
To simultaneously collect interaction data across diverse object types and achieve stable object-robot interactions, we build a hierarchical RL training pipeline to train a planar velocity controller for various objects, as illustrated in Fig.~\ref{fig:1}(b). The pipeline includes four main components: a low-level WBC, a high-level object controller, task-specific reward design, and multi-object environment generation.

\subsection{Low-Level Whole-Body Controller}
The low-level controller, denoted as $\pi^{\text{low}}: \mathcal{N} \times \mathcal{A} \times \mathcal{Z} \rightarrow \mathcal{T}$, generates target joint positions $\mathbf{q}_t^{\text{des}} \in \mathbb{R}^{18}$ of the legged manipulator, and converts it to the joint torque using the proportional-derivative control law. Here, $\mathcal{N}$ represents the proprioceptive states, comprising joint positions $\mathbf{q}_t \in \mathbb{R}^{18}$, joint velocities $\mathbf{\dot{q}}_t \in \mathbb{R}^{18}$, base orientation (row, pitch) and angular velocities $\mathbf{s}^{\text{base}}_t \in \mathbb{R}^5$. The action space $\mathcal{A} $ denotes the base velocity command and end-effector pose. Besides, the low-level controller employs a convolutional neural network (CNN) to encode the history of proprioceptive states into a latent variable $\mathbf{z}_t \in \mathbb{R}^{18}$, which encodes temporal observations for coordinated whole-body control. In summary, the command of the low-level control policy is:
\begin{equation}
    \mathbf{c}_t^{\text{low}} = [\mathbf{p}_{\text{ee}}, \mathbf{q}_{\text{ee}}, v_x^r, v_y^r, \omega_z^r],
\end{equation}
where $\mathbf{p}_{\text{ee}} \in \mathbb{R}^3$ and $\mathbf{q}_{\text{ee}} \in \mathbb{R}^3$ are position and orientation of end-effector in robot's frame, $ v_x^r$, $v_y^r$ are the robot's linear velocity, and $\omega_z^r$ denotes the angular velocity. The observation of the low-level policy is:
\begin{equation}
    \mathbf{o}_t^{\text{low}} = [\mathbf{q}_t, \mathbf{\dot{q}}_t, \mathbf{s}^{\text{base}}_t, \mathbf{a}_{t-1}^{\text{low}}, \mathbf{z}_t, \mathbf{c}_t^{\text{low}}, \mathbf{t}_t],
\end{equation}
where $\mathbf{t}_t \in \mathbb{R}^5$ is gait parameters, $\mathbf{a}_{t-1}^{\text{low}} \in \mathbb{R}^{12}$ is last action of the low-level policy. The RL-based neural network architecture and reward functions are inspired by \cite{liu2024visual}. As illustrated in Fig. \ref{fig:1}(b), the low-level controller's role is to track the action output of the high-level controller $\mathbf{a}_t^{\text{high}}$, providing the foundation that enables the high-level policy to perform versatile manipulation across different objects.

\subsection{High-Level Object Controller}
As shown in Fig. \ref{fig:1}(b), training on top of the low-level WBC, the high-level object velocity controller mainly consists of three elements: the ICR module $h_l: \mathcal{N} \times \mathcal{B} \times \mathcal{R} \rightarrow \mathcal{L}$, the object velocity estimator $h_{v}: \mathcal{N} \times \mathcal{R} \rightarrow \mathcal{V}$, and the high-level policy $\pi^{\text{high}}: \mathcal{N} \times \mathcal{V} \times \mathcal{B} \times \mathcal{L} \rightarrow \mathcal{A}$, where $\mathcal{B}$ denotes the object states, $\mathcal{R}$ indicates the privileged states, $\mathcal{L}$ is the latent representation of ICR module, and $\mathcal{V}$ is the predicted object's velocity. The action space $\mathcal{A}$ indicates the robot's velocity and the increment of six joint angles of the manipulator $\mathbf{a}^{\text{high}}_t \in \mathbb{R}^{9}$. The command of the high-level policy is the object velocity:
\begin{equation}
    \mathbf{c}_t^{\text{high}} = [ v_x^b, v_y^b, \omega_z^b].
\end{equation}
Unlike the low-level policy, the observation of the high-level policy consists of the robot’s proprioceptive states and the object states:
\begin{equation}
    \mathbf{o}_t^{\text{high}} = [\mathbf{q}_t, \mathbf{\dot{q}}_t, \mathbf{s}^{\text{base}}_t, \mathbf{a}_{t-1}^{\text{high}}, \mathbf{c}_t^{\text{high}}, \mathbf{p}_{\text{ee}}, \mathbf{q}_{\text{ee}}, \mathbf{p}_{\text{obj}}, \mathbf{q}_{\text{obj}}, \mathbf{\hat{v}}^b,\mathbf{l}_t, ],
\end{equation}
where $\mathbf{p}_{\text{obj}} \in \mathbb{R}^3, \mathbf{q}_{\text{obj}} \in \mathbb{R}^3$ are the position and orientation of the object in the robot frame. $\mathbf{l}_t \in \mathbb{R}^{128}$ and $\mathbf{\hat{v}}^b \in \mathbb{R}^3$  are the encoded graph-level features and the estimated object's velocity, which are introduced in Sec. \ref{sec:icr} and Sec. \ref{sec:vel}.

During training,  we concurrently train the ICR module and the object velocity estimator with the high-level policy using end-to-end model-free RL \cite{rudin2022learning}. The low-level WBC policy is pretrained and fixed. 
In addition to the predicted velocity of the object $\hat{\mathbf{v}}^b$, the embedding feature of interactive configuration $\mathbf{l}_t$, the high-level policy $\pi^{\text{high}}$ also takes as input the past 10 observation frames $(\mathbf{o^{\text{high}}_{t-9}}, \cdots, \mathbf{o^{\text{high}}_{t-1}}, \mathbf{o^{\text{high}}_{t}})$ to predict the action $\mathbf{a}_t^{\text{high}}$. The predicted high-level action $\mathbf{a}_t^{\text{high}}$ is then fed into the low-level controller $\pi^{\text{low}}$ to track the command of each joint. In general, the RL objective is to maximize the accumulated discounted reward, which is defined as:
\begin{equation}
J(\pi^{\text{high}}, \pi^{\text{low}}, h_{v}, h_l) = \mathbb{E}_{{\tau} \sim p(\tau \mid \pi^{\text{high}}, \pi^{\text{low}}, h_{v}, h_l)} 
\left[ \sum_{t=0}^{T-1} \gamma r_{t} \right],
\end{equation}
where $r_t$ is the immediate reward received at time step $t$, $\gamma \in [0, 1]$  is the discounted factor, and $T$ is the trajectory horizon (episode length). The entire hierarchical controller training procedure is shown in Algorithm \ref{alg1}.

\begin{algorithm}[t]
\label{alg1}
\caption{Hierarchical Controller Training.}
\For{$\textbf{episode}=1 \ \KwTo \ M$}{
  Initialize $\pi^{\text{low}}, h_{v},\, h_{l}$\;
  Initialize proprioceptive states history $H_{\text{pro}}$\;
  Initialize privileged states history $H_{\text{pri}}$\;
  Initialize object states history $H_{\text{obj}}$\;
  Initialize data batch $D$\;
  Sample velocity command $\mathbf{v}^b$
  
 \CommentLine{High-Level Controller, 50 Hz} % 整行注释
  \For{$t=1 \ \KwTo \ T$}{
    $\mathbf{\hat{v}}^{b} = h_{v}(H_{\text{pro}}, H_{\text{pri}})$ \tcp*[r]{\textcolor{blue}{\text{Object Velocity Estimator}}}
    $G \leftarrow \text{Building IG} (H_{\text{pro}}, H_{\text{obj}})$
    
    $\mathbf{l}_{t} = h_{l}(G)$ \tcp*[r]{\textcolor{blue}{\text{ICR Module}}}
        
    $\mathbf{a}_t^{\text{high}}\sim \pi^{\text{high}}_\theta\!\left(\mathbf{a}_t \mid \mathbf{l}_{t},\,\mathbf{\hat{v}}^b, H_{\text{pro}}, H_{\text{obj}}\right)$\;
        
    \CommentLine{Low-Level Controller, 200 Hz}
    \For{$t^{\prime}=1 \ \KwTo \ K$}{
    $\tau^{\text{robot}}_{t^{\prime}} = \pi^{\text{low}}\!\left( \mathbf{a}_{t}, H_{\text{pro}} \right)$ \tcp*[r]{\textcolor{blue}{\text{Fixed $\pi^{\text{low}}$}}}
      Execute $\tau^{\text{robot}}_{t^{\prime}}$\;
    }
    Store transition $\mathbf{s}_{t},\mathbf{a}_{t}, \mathbf{r}_t$ in $D$\;
  }

  $L_{\text{estimator}} = \left\| \hat{\mathbf{v}}^{b} - \mathbf{v}^{b} \right\|_2^2$
   
  $L_{\text{total}} = \mathbb{E}_{{\tau} \sim p(\tau \mid \pi^{\text{high}}, \pi^{\text{low}}, h_{v}, h_l)} \left[ \sum_{t=0}^{T} \gamma^{t} r_{t} \right]$

  Compute $\nabla_{v} L_{\text{estimator}}$ 

  Compute $\nabla_{\theta,l} L_{\text{total}}$, \tcp*[r]{\textcolor{blue}{\text{PPO \cite{rudin2022learning}}}}
  
  Update $\pi_{\theta}^{\text{high}}, h_v, h_l$
}
\end{algorithm}

\subsection{Task-Specific Reward Functions}
Compared with manipulating lightweight and small items, rearranging large objects requires careful consideration of safety and object stability. First, the legged manipulator must avoid collisions with the object during manipulation to ensure safe and stable interaction.
Second, the manipulator’s joint torque limits must be considered to prevent overload and mechanical damage.
\begin{table}[t]
    \centering
    \caption{Reward Functions}
\begin{tabular}{cccc}
    \toprule
    \textbf{Term}& \textbf{Reward} & \textbf{Expression}& \textbf{Weight} \\
     \midrule
     \multirow{2}{*}{Task} 
       & Lin. Vel. Tracking &  $e ^{-4\|\mathbf{v}_{xy}^* - \mathbf{v}_{xy}\|^2}$ & 5.0 \\

      & Ang. Vel. Tracking &  $e ^{-4(\mathbf{\omega}_{\text{yaw}}^* - \mathbf{\omega}_{\text{yaw}})^2}$  & 5.0  \\

     \midrule

       \multirow{3}{*}{Collision} 
       & Distance Keeping  &  $\frac{1}{1 + e^ {200 (d_x - d_{\text{th}}) }}$ & -10.0 \\

       & Yaw Alignment & $  - \frac{\left| \psi_o - \psi_r \right|}{\pi} $ & 5.0  \\

       & Contact Penalize & $\sum_{i\in \mathcal{I}}\mathbb{I}\!\left( \lVert \mathbf{F}_{i}\rVert_2 > \hat{\tau} \right)$ & -5.0 \\
       
     \midrule

     \multirow{3}{*}{Effort} 
       & Joint Torque &  $\| \boldsymbol{\tau} \|^2$ & $-2.5e^{-5}$ \\
       & Joint Acc &  $\| \mathbf{\ddot{q}} \|^2$ & $-2.5e^{-7}$ \\
       & EE Wrench &  $\| \mathbf{w}_{\text{ee}} \|^2$ & $1.0e^{-3}$ \\

    \midrule

    \multirow{3}{*}{Smoothness} 
       & Action Rate &  $\|\mathbf{a}_t - 2\mathbf{a}_{t-1} + \mathbf{a}_{t-2} \|^2$ & $-2.0e^{-3}$ \\
       & Lin. Vel.($z$)&  $\mathbf{v}_z^2$ & -2.0 \\
       & Ang. Vel.($xy$) &  $\| \boldsymbol{\omega}_{xy} \|^2$ & -0.05\\
       
    \midrule
    
    \multirow{2}{*}{Pose} 
       & Orientation &  $\| \mathbf{g}_{xy} \|^2$ & -10.0 \\
       & Default Joint &  $\sum_{j=0}^{5} (q_j - q^\star_j)^2$  & -1.0 \\
     \bottomrule
\end{tabular}
    \label{tab:reward}
\end{table}
Third, the object’s pose must remain stable throughout manipulation to prevent toppling and potential damage. This requires adaptive control strategies that optimize both the joint positions and apply force, which is similar to how humans intuitively adjust their arm posture when pushing objects. We design tailored reward functions to address the problems above. Specifically, the self-collision avoidance reward encourages the robot to maintain the object within its frontal workspace while avoiding excessive yaw misalignment. It is formulated as:
\begin{equation}\label{eq:collision}
\begin{aligned}
    r_{t}^{\text{collision}} = & c_1 \cdot \left( - \frac{\left| \psi_o - \psi_r \right|}{\pi} \right) +  c_2 \cdot \frac{1}{1 + e^{ 200 (d_x - d_{\text{th}})}} \\
    & +  c_3 \cdot \sum_{i\in \mathcal{I}} \mathbb{I}\!\left( \lVert \mathbf{F}_{i}\rVert_2 > \hat{\tau} \right),
\end{aligned}
\end{equation}
where $\psi_o$ and $\psi_r$ denote the yaw angles of the object and the robot, respectively, $d_x$ is the distance between end-effector and robot base along the $x$-axis, $d_{\text{th}}$ is a safe distance threshold, and $200$ is a scaling factor controlling the steepness of the penalty. $\mathbb{I}(\cdot)$ is the indicator function, which returns 1 if the condition is true and 0 otherwise, and $\hat{\tau} = 1.0$. $\mathbf{F}_{i}$ denotes the force vector along $x,y,z$ applied on the $i$ th undesired-contact body, which includes the base, calf, and thigh joints.  
Besides, to limit excessive manipulator efforts and avoid hardware damage, we design the hardware-safe actuation penalty term, which penalizes large joint torques, end-effector forces, and abrupt joint accelerations:  
\begin{equation}\label{eq:effort}
\begin{aligned}
    r_t^{\text{effort}} &= c_4 \cdot \sum_{j = 1}^6 \tau_j^2 + c_5 \left\lVert \mathbf{w}_{\text{ee}} \right\rVert_2^2 + c_6 \cdot \sum_{j=1}^{6} \ddot{q}_j^2,
\end{aligned}
\end{equation}
where $\tau_j$ is the torque at the $j$-th arm joint, $\mathbf{w}_{\text{ee}} \in \mathbb{R}^6$ is the six-dimensional wrench (three forces and three moments) measured at the end-effector, $\ddot{q}_j$ denotes the angular acceleration of the $j$-th joint, and $c_i$ is the hyperparameter. Furthermore, real-time observations of the object pose are incorporated into the policy input, and object pose stability terms are imposed on variations in the object’s roll and pitch angles.
This design promotes stable object control and effectively prevents object toppling during manipulation. 
In addition to the specially designed rewards, Table~\ref{tab:reward} illustrates several standard reward terms incorporated during RL training, including linear and angular velocity tracking rewards, smoothness rewards, similar to those used in \cite{11196002}.

\subsection{Multi-Object Environment Generation}
To collect interaction data between the robot and multiple objects within a single training process, we build a scalable multi-object training environment that supports $M$ object categories ($M>0$). Specifically, given a total of $Q$ parallel environments, we allocate $\lfloor Q/M \rfloor$ instances to each object category. During training, interaction data from all environments are aggregated into a unified replay buffer and jointly used to optimize the high-level policy, thus enabling sample-efficient learning of interactions across diverse objects.

\begin{figure*}[t]	
	\centering
        % \vspace{3pt}
	\includegraphics[width=0.98\linewidth]{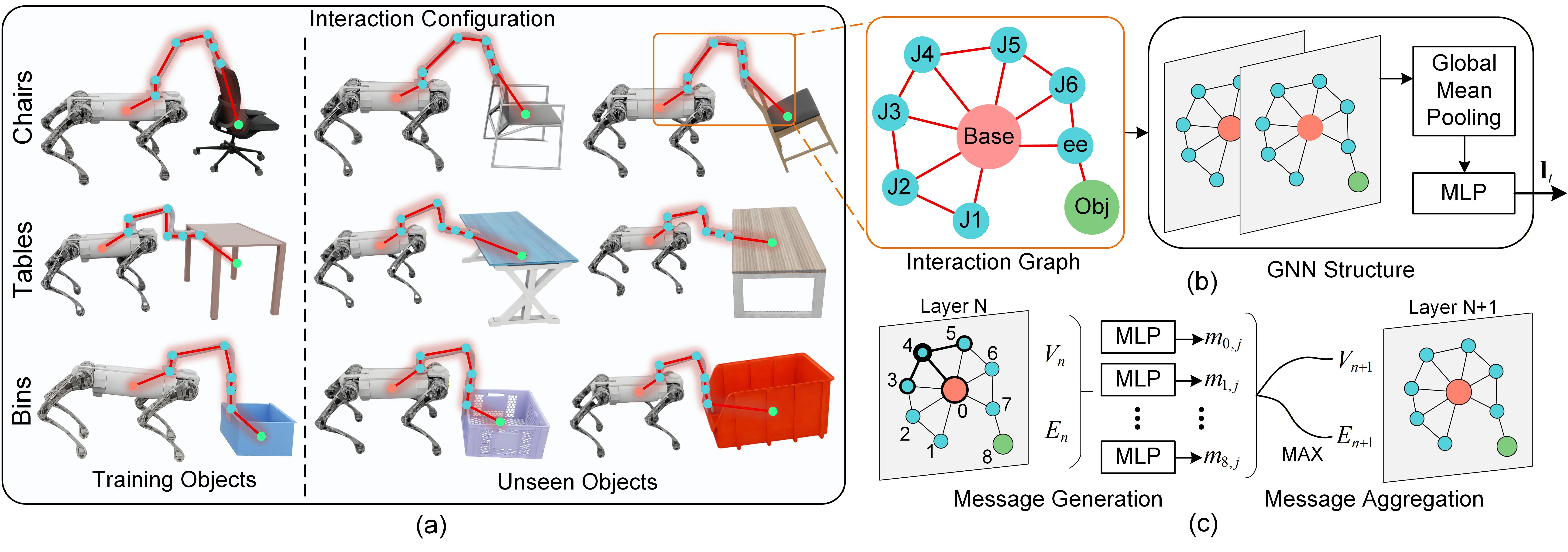}
	\setlength{\abovecaptionskip}{-1pt} 
	\caption{Illustration of interaction configuration representation. (a) Interaction configuration examples when the robot interacts with three different types of objects. The red line segments represent the topology of the interaction configuration, while the pink, blue, and green nodes correspond to the nodes on the robot, the manipulator, and the objects, respectively. (b) The robot-object interactive configuration is modeled as an IG, where each node or edge contains the features of the robot and object. The graph is then processed by a GNN to extract the graph-level feature. (c) Details of the message generation and aggregation process during GNN training.
    }
	\label{fig:3}
\end{figure*}

Without loss of generality, we train on three distinct object categories in this work (i.e., $M=3$): a chair ($0.5\,\mathrm{m}\times0.5\,\mathrm{m}\times0.8\,\mathrm{m}$), a table ($1.0\,\mathrm{m}\times0.5\,\mathrm{m}\times0.6\,\mathrm{m}$), and a bin ($0.6\,\mathrm{m}\times0.4\,\mathrm{m}\times0.3\,\mathrm{m}$), as shown in the first column of Fig.~\ref{fig:3}(a). 
During training, we employ 900 parallel environments, allocating 300 instances to each object type to collect interaction data efficiently. 
At the beginning of each episode, the manipulator is initialized at the designated grasp point, consistent with the initialization protocol in \cite{li2025robotmover}. This setup allows us to focus on post-grasp object velocity control, while grasping execution and pose estimation are delegated to other subsystems. To improve sim-to-real robustness, we further apply domain randomization over three categories of physical parameters: grasp-point perturbations, the object-ground friction coefficient, and the object mass (Table~\ref{tab:randomize}). Notably, the grasp-point randomization range is object-dependent and is constrained by each object's feasible graspable region. 
An episode is terminated and the environment is reset if the object roll or pitch exceeds $0.8$~rad, or if the gripper loses contact with the object.

\begin{table}[t]
    \centering
    \caption{Parameters Randomization}
    \resizebox{0.85\columnwidth}{!}{%
\begin{tabular}{ccc}
    \toprule
    \textbf{Parameters} & \textbf{Range} & \textbf{Unit} \\
     \midrule
     
     \multirow{3}{*}{Grasp Position} 
     & Chair: [-10, 10] &  cm   \\
     & Table: [-20, 20] &  cm   \\
     & bin: [-15, 15] &    cm   \\
     \midrule
     
      Friction Coefficient & [0.1, 0.6] &  -  \\
      \midrule

    Object Mass & [5, 15] &  kg    \\
    \midrule

    Object Lin. Vel. Command & [-0.5, 0.5] & m/s \\
    \midrule

    Object Ang. Vel. Command & [-0.5, 0.5] & rad/s \\
       
     \bottomrule
\end{tabular}
}
    \label{tab:randomize}
\end{table}

\section{Learning  Velocity Control Policy for Various Objects}
After establishing an RL training pipeline that supports interaction data collection across diverse objects, we next introduce two carefully designed modules, the ICR module and the object velocity estimator, which improve the generalization of a single policy, enabling it to better adapt to different objects.
% we next consider how to leverage these data to improve the velocity tracking accuracy of a single policy when controlling different objects. 
% Fundamentally, rigid-body motion is mainly determined by two factors: (i) the external wrench, and (ii) the object’s physical properties, such as mass, inertia, and the friction coefficient. In practice, the physical parameters are difficult to reliably predict \cite{song2020probabilistic} or identify online \cite{zhang2025provably}. Moreover, for the same joint torques, the induced object wrench changes with robot-object interaction configuration, because both the Jacobian-based actuation-to-wrench transmission and the contact wrench mapping (lever arms) are configuration-dependent. In other words, under different interaction configurations, applying the same wrench to the object typically requires different action distributions.
% Based on this analysis, our key idea is to encode the robot-object interaction configuration into a latent embedding conditioned on the policy to produce configuration-dependent actions, thereby stabilizing control across different objects. 
% Moreover, we explicitly estimate the object’s velocity and provide closed-loop feedback to further improve velocity tracking accuracy. 

\subsection{Interaction Configuration Representation} \label{sec:icr}
% We aim to accurately control the planar velocity of diverse objects using a single policy. However, a key challenge lies in the fact that the action distributions—particularly for arm motions—can vary significantly when interacting with different objects. Essentially, we argue that such variations arise primarily from differences in robot-object interaction configurations, rather than being directly determined by explicit object type or geometry. Because identical robot actions can induce different object wrenches under different interaction configurations, leading to distinct object motions. For example, pushing a square or a round table from similar contact configurations leads to comparable control behaviors, as the relative contact location dominates force transmission and coordination patterns instead of the precise object shape. Motivated by this observation, instead of explicitly identifying object types, 
% As mentioned above, we aim to implicitly encode interaction configurations to condition the high-level policy, enabling it to generate distinct action distributions across different objects.
% To this end, 
Inspired by graph-based methods for physical interaction \cite{zhang2023simulation}, we propose a unified representation for robot-object interaction, the ICR module, that enables a legged manipulator to regulate the planar velocity of diverse objects via pushing or pulling using a single control policy. The overall process is illustrated in Fig. \ref{fig:3}. Specifically, this module leverages the graph structure, with the quadruped, robot arm, and object as the nodes, and their spatial relationships (e.g., relative pose) as the edges to represent the overall interactive state, which is referred to as an IG. Then, we leverage GNNs, a powerful tool that captures relational structures in graphs \cite{lai2025roboballet}, to embed the interaction graph into a graph-level feature $\mathbf{l}_t$, which is fed to the high-level policy to produce configuration-adaptive actions. Here, the intuition is that different interaction configurations require different action distributions for stable object interaction. Meanwhile, the main advantage of using GNNs for feature extraction over simply flattening the state information into a vector, as done in CNNs, is that it provides a reusable way to incorporate relational information between robots and objects, allowing the network to reason about them based on their actual physical relationships \cite{roboball}. The proposed ICR module consists of two stages: IG construction and graph-level feature learning.

\subsubsection{Interaction Graph Construction} To model the robot-object interaction configuration, we build an IG, denoted by $G = (N, E)$, as shown in Fig. \ref{fig:3}(b). This graph is constructed according to the actual physical couplings: the quadruped and the manipulator within the legged-manipulator system are directly coupled and mutually influence each other, while the object is connected to the system via the end-effector. Thus, all six joints of the manipulator are connected to the quadruped, the arm joints are linked in series, and the end-effector is connected to the object. All edges are modeled as bidirectional, facilitating efficient information sharing across the entire graph. Specifically, the built graph contains nodes $n_0, \cdots, n_8$, representing four types of entities: quadruped, six arm joints, the end-effector, and the manipulated object. Each node $n_i$ is associated with a feature vector $\mathbf{f}_{n_i}$: 
\begin{itemize}
    \item Quadruped base node ($n_0$): $\mathbf{f}_{n_0} = [\boldsymbol{\phi}_b, \boldsymbol{\omega}_b] \in \mathbb{R}^5$, where $\boldsymbol{\phi}_b \in \mathbb{R}^2$ is the planar orientation and $\boldsymbol{\omega}_b \in \mathbb{R}^3$ is the angular velocity. 
    \item Joint node $(n_i, i=1,\cdots,6)$: 
    $\mathbf{f}_{n_i} = [\mathbf{r}_i, \; q_i, \; q^{\mathrm{def}}_i,  (q_i - q^{\mathrm{def}}_i), \; \dot{q}_i]\ \in \mathbb{R}^{11}$, 
    where \(\mathbf{r}_i \in \mathbb{R}^7\) is the relative pose of link $i$ to the robot base, $q_i$ is the angle of joint $i$, and $q_i^{\text{def}}$ is the default angle of joint $i$.
    \item End-effector node (\(n_7\)): 
    $\mathbf{f}_{n_7} = [\mathbf{r}_{\text{ee}}, c_{\text{ee}}] \in \mathbb{R}^8$, 
    where \(\mathbf{r}_{\text{ee}} \in \mathbb{R}^7\) is the relative pose to the quadruped base and \(c_{\text{ee}} \) is the contact state, $c_{\text{ee}} = 1$ if the end-effector is in contact with the object, and 0 otherwise.
    \item Object node (\(n_8\)): 
    $\mathbf{f}_{n_8} = [\mathbf{r}_o, v_x^b, v_y^b, \omega_z^b] \in \mathbb{R}^{10}$, where \(\mathbf{r}_o \in \mathbb{R}^7\) is the object pose (geometric center) in the quadruped base frame and $(v_x^b, v_y^b, \omega_z^b)$ are the object velocity command.
\end{itemize}
All feature vectors are zero-padded to $\mathbb{R}^{11}$ and concatenated with a one-hot type vector of length 4. So, finally, each node's feature vector $\mathbf{f}_{n_i} \in \mathbb{R}^{15}$. Additionally, the edge set $E$ is defined as:

\begin{equation}
\begin{aligned}
    E =  &\{e_{0,j} \mid j=1,\cdots,7\} \\
  &\cup\ \{e_{j,j+1} \mid j=1,\cdots,6\}  \\
  &\cup\ \{e_{7,8}\},
\end{aligned}
\end{equation}
where $e_{0,j} (j=1,\cdots,7)$ denotes edges connecting the quadruped base node to each arm joint node, $e_{j,j+1} (j=1,\cdots,6)$ represents edges along the arm joint chain, and $e_{7,8}$ connects the end-effector to the object. Each feature vector $\mathbf{f}_{e_{i,j}}$ of each edge $e_{i,j}$ is defined as:
\begin{equation}
    \mathbf{f}_{e_{i,j}} = [\mathbf{p}_i - \mathbf{p}_j, \mathbf{q}_i \otimes \mathbf{q}_j^{-1}] \in \mathbb{R}^7,
\end{equation}
which is the relative position and orientation between connected nodes $n_i$ and $n_j$. In this way, the built IG contains both the topological structure and dynamic states of the robot-object interaction system. Next, we introduce how to extract the graph-level features using GNNs for high-level policy learning.

\subsubsection{Graph-Level Feature Learning} Given the IG, our goal is to predict the property of an entire graph, which captures different interaction configurations between the robot and objects. As illustrated in Fig. \ref{fig:3}(b), the feature extraction process consists of three steps. First, the graph is processed through two GNN layers, each consisting of an EdgeConv operator for message passing and aggregation \cite{wang2019dynamic}. Then, a global mean pooling operation is applied to obtain the graph-level representation. Finally, the resulting vector is passed through an MLP to map it to the final feature $\mathbf{l}_t \in \mathbb{R}^{128}$, which serves as the input to the high-level policy. In the following, we detail the message generation and aggregation process within the GNN layers, the core stages of the learning procedure.

As shown in Fig. \ref{fig:3}(c), the message generation process transforms node and edge features into interaction-aware messages through MLPs. For each edge $e_{i,j}$, the message $\mathbf{m}_{i,j}$ is computed as:
\begin{equation}
    \mathbf{m}_{i,j}^l = \text{MLP}^{(l)}(\mathbf{h}_i^l, \mathbf{h}_j^l, \mathbf{f}_{e_{i,j}}),
\end{equation}
where $\mathbf{h}_i^l$ and $\mathbf{h}_j^l$  are hidden features of target and source nodes at layer $l$, $\mathbf{f}_{e_{i,j}}$ is the edge attributes. $\mathbf{m}_{i,j}^l$ is the message propagated from node $j$ to node $i$, where node $j$ is one of the neighbors of node $i$. It is worth noting that node $i$ might have more than one neighbor. For example, as indicated by the bold black lines in Fig. \ref{fig:3}(c), the neighbors of $n_4$ are $n_3, n_5$, and $n_0$.

After the message generation, the message aggregation step is to update the node feature utilizing the gathered message $\mathbf{m}_{i,j}$ from neighbors. For each node $n_i$, the feature can be updated by:
\begin{equation}
    \mathbf{h}_i^{l+1} = \rho \left(\mathbf{m}_{i,j}^l \ | \  j\in N_{in}(i) \right),
\end{equation}
where $N_{in}(i)$ denotes the set of all incoming neighbors of node $i$, and $\rho(\cdot)$ is instantiated as the max function in our setting to capture the most related feature. After completing message aggregation, the node features of the next GNN layer are updated.

\subsection{Object Velocity Estimator} \label{sec:vel} 
For the object velocity estimator $h_{v}$, we design a supervised learning module that predicts the various objects' planar velocity. The estimator takes as input a history of proprioceptive states over the past $10$ time steps, which encapsulates the robot's joint and base kinematics while interacting with the object. This temporal sequence is first encoded using a long short-term memory (LSTM) network, followed by a multilayer perceptron (MLP) that outputs the estimated object velocity $\hat{\mathbf{v}}^{b} = [\hat{v}_x^{b}, \hat{v}_y^{b}, \hat{\omega}^{b}_z]$. The network parameters are optimized by minimizing the mean squared error (MSE) between the predicted velocity $\hat{\mathbf{v}}^{b}$ and the ground-truth velocity $\mathbf{v}^{b}$:
\begin{equation}
J_{v} = \left\| \hat{\mathbf{v}}^{b} - \mathbf{v}^{b} \right\|_2^2 ,
\end{equation}
where the ground-truth object velocity $\mathbf{v}^{b}$ is obtained from privileged information during training. This estimator enables the high-level policy to receive closed-loop feedback of the object's velocity without relying on external sensors, thus enhancing the accuracy of object velocity control.

\section{Task and Motion Planning For Multi-Object Rearrangement}\label{sec:TAMP}
Once we can accurately control an object’s planar velocity, large-object rearrangement can be decomposed into collision-free object trajectory generation and trajectory tracking. In this section, we present a task planner that generates efficient multi-object rearrangement sequences, including the object visitation order and the object-to-target assignment. Meanwhile, we further develop a coarse-to-fine robot trajectory planner that supports online replanning for accurate object grasping.

\subsection{Task Planning with Greedy Warm Start}
The goal of task planning is to optimize discrete sequences of object grasp and release to minimize overall task completion time. To model this task, we adopt the symbolic representation commonly used in the TAMP \cite{10705419}. Let the robot's configuration be denoted by $ \mathcal{C}^r$, which contains the robot's global position, joint states, and end-effector status. The robot’s full state is represented as $\mathbf{s}^r = ( \mathcal{C}^r, h^r)$, where $h^r \in \{\text{empty}\} \cup \mathcal{O}$ indicates whether the robot is currently holding an object. The environment contains a set of $N$ target objects and a corresponding set of $N$ target locations, denoted by $[ to_1, \cdots, to_N ]$ and $[ tl_1, \cdots, tl_N ]$, respectively. The robot can execute two primitive manipulation actions: \textit{Grasp}, which is feasible when the robot is not holding any object ($h^r = \text{empty}$) and its configuration satisfies $\mathcal{C}^r \in \text{Reach}(to_i)$; and \textit{Release}, which is applicable when $h^r = to_i$ and $\mathcal{C}^r \in \text{Reach}(tl_j)$. A complete task plan can be represented by an object visiting order $\boldsymbol{\phi} = [\phi_1, \cdots, \phi_N]$ and an object-to-target assignment $\boldsymbol{\sigma} = [\sigma_1, \cdots, \sigma_N]$, together with a sequence of continuous trajectories $ [\tau_1, \cdots, \tau_N]$. To find a globally optimal plan, we formulate an optimization problem as:
\begin{equation}
\label{eq:opt_problem}
\begin{aligned}
\min_{\phi, \sigma, \tau_k} \quad & \sum_{k=1}^{N} C(\tau_k) \\
\text{s.t.} \quad
& \boldsymbol{\phi} \in \mathbb{S}_N, \quad \boldsymbol{\sigma} \in \mathbb{B}_N, \\
& \tau_k : \mathcal{C}_{k-1}^r \xrightarrow{} to_{\phi_k} \xrightarrow{} tl_{\sigma_{k}}
, \\
& \tau_k \subset \mathcal{C}_{\text{free}}, \\
& \mathcal{C}_{k-1}^r \in \text{Reach}(to_{\phi_k}), \quad
  tl_{\sigma_{k}} \in \text{Reach}(to_{\phi_k}), 
\end{aligned}
\end{equation}
where $C(\tau_k)$ denotes the execution time of trajectory $\tau_k$, $\phi \in \mathbb{S}_N$ denotes a permutation over the set of object indices. $\sigma \in \mathbb{B}_N$ represents a bijective assignment from objects to target locations. The trajectory $\tau_k$ corresponds to the continuous motion for completing the $k$-th grasp-and-release operation. It can be decomposed into two segments: (i) A trajectory $\tau_k^{\text{pre}}$ that moves the robot from its previous configuration $\mathcal{C}_{k-1}^r$ to a graspable pose near object $o_{\phi_k}$, and (ii) A trajectory $\tau_k^{\text{post}}$ that moves the robot, while holding the object, from the grasp location to the assigned target $tl_{\sigma_{k}}$. Both trajectories must be collision-free and satisfy the robot’s kinematic and reachability constraints. The generation details of  $\tau_k^{\text{pre}}$ and $\tau_k^{\text{post}}$ are introduced in Sec. \ref{sec:Traj}. Additionally, the pre-grasp configuration $\mathcal{C}_{k-1}^r$ and the placement target $tl_{\sigma_{k}}$ should both be reachable given the geometry and kinematic constraints of the robot, as captured by the reachability condition $\text{Reach}(\cdot)$.

To solve the problem (\ref{eq:opt_problem}), we adopt a hybrid Branch and Bound (BnB) algorithm \cite{7526562} enhanced with an added greedy warm start. The algorithm first computes a greedy solution to initialize a lower bound of the problem (\ref{eq:opt_problem}). Subsequently, the Hungarian algorithm and Prim’s algorithm for minimum spanning tree \cite{parker2016multiple} are employed to compute a tight lower bound iteratively. Compared to enumeration or heuristic approaches, this method reduces computation time by effective pruning of non-promising branches, and the BnB framework inherently guarantees global optimality \cite{7526562}.

\subsection{Coarse-to-Fine Trajectory Generation}\label{sec:Traj}
To generate trajectories $\tau_k^{\text{pre}}$ and $\tau_k^{\text{post}}$, we model the robot in the pre-grasp stage and the grasped object in the post-grasp stage using the differential-drive kinematic formulation on $\mathrm{SE}(2)$:
\begin{equation}
\dot{x} = v_x \cos\theta, \quad  
\dot{y} = v_x \sin\theta, \quad  
\dot{\theta} = \omega_z,
\end{equation}
with control inputs $v_x$ and $\omega_z$. 
Based on this kinematic model, 
We extend the motion state-based polynomial optimization approach \cite{10924228}, which is originally developed for differential-drive robots, for multi-object planning. Specifically, we incorporate the task planning module into \cite{10924228} and upgrade the original single-stage planner to a two-stage coarse-to-fine scheme to support accurate object grasping. For the trajectory $\tau_k^{\text{pre}}$, it follows a coarse-to-fine planning method. Since the object’s initial pose is only approximately known, we first generate a coarse approach trajectory that moves the robot into the object’s vicinity. Once the robot is within a 5-meter range of the estimated object location, target detection is triggered. Upon detecting the specific object, the perception module evaluates the relative position and orientation between the object and the robot. Based on the detected object poses, we replan a more precise trajectory so that the robot can reach the designated grasping pose.

To generate safety trajectories, we approximate the robot-object system as a set of collision circles. For $\tau_k^{\text{pre}}$, the collision model comprises three circles covering the quadruped robot and the manipulator. For $\tau_k^{\text{post}}$, the model is extended to represent the entire robot-object system by additionally introducing a collision circle centered at the object's geometric center. Considering that the manipulator adaptively adjusts its joint angles within a designed range during object arrangement. Thus, although this collision model is conservative, it ensures safe execution under such joint-adaptive interaction. 

Moreover, throughout the object rearrangement process, the map is dynamically updated with respect to the currently manipulated object, ensuring collision avoidance with other objects. For example, when generating $\tau_k^{\text{pre}}$, the object's current position is treated as an obstacle to prevent collision. After the object is grasped, its position is updated to free space, while other objects remain as obstacles.

\section{Experiments and Results} \label{sec:exp}
In this section, we validate the proposed key components and entire system through extensive simulations and real-world experiments. Currently, to the best of the authors’ knowledge, no prior method accomplishes a complete system for large-object rearrangement that integrates perception, grasping, and long-horizon object rearrangement; therefore, for comparison, Sec. \ref{subsec:Comparison} focuses on object control performance after grasping. Specifically, we compare the proposed high-level object controller with state-of-the-art (SOTA) baselines in terms of average training reward, velocity-tracking accuracy, reference-tracking performance, as well as qualitative comparisons with other baselines using two intuitive examples.
In Sec. \ref{subsec:ICR}, we analyze two key components by visualizing the ICR module and the object-velocity estimator. In Sec. \ref{subsec:unseenObj}, we evaluate the generalization capability of the high-level policy through velocity tracking experiments on six previously unseen objects. Besides, in Sec.~\ref{sec:eva_task_planning}, we compare the proposed task planning method against a greedy baseline in terms of completion time and total travel distance. Moreover, in Sec. \ref{subsec:simexp} and \ref{subsec:realexp}, we present comprehensive object rearrangement experiments across multiple objects and scenarios in both simulation and real-world settings, which demonstrate the practicality and strong robustness of the proposed large-object rearrangement system.

\begin{figure}[t]	
	\centering
	\includegraphics[width=0.93\linewidth]{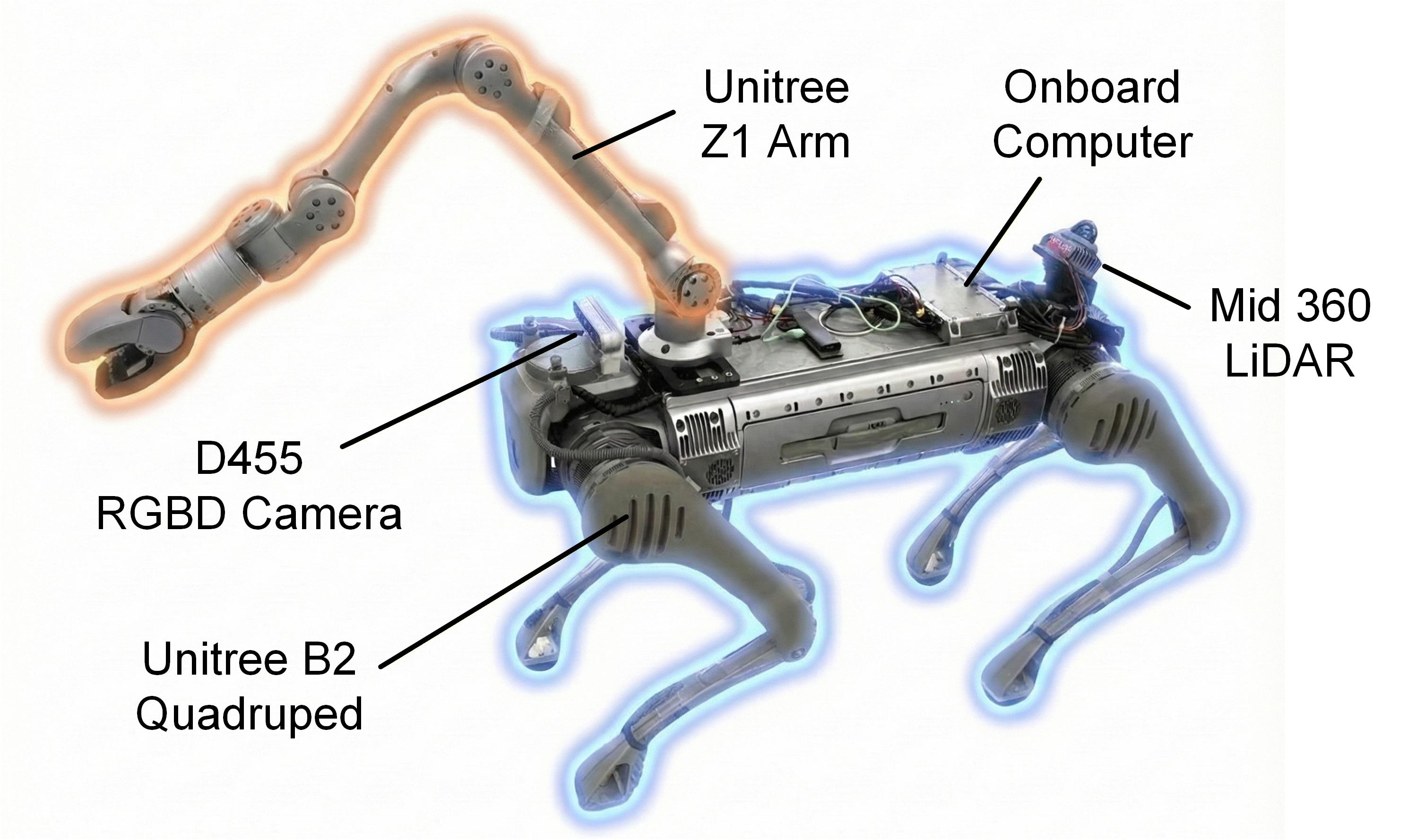}
	\setlength{\abovecaptionskip}{-1pt} 
	\caption{System setup of the legged manipulator.}
	\label{fig:11}
    \vspace{-5pt}
\end{figure}

\subsection{Experimental Setup} \label{sec:exp_set_up}
\subsubsection{Simulation Settings}
All training is conducted on a single workstation equipped with an Intel Core i9 3.60 GHz processor and a GeForce RTX 4080 GPU. It takes about 6 hours to train 900 environments and 10000 iterations parallel on the Isaac Sim simulator \cite{mittal2023orbit}. The high-level policy operates at a frequency of 50 Hz, with a maximum episode length of 20 seconds. During training, we use the manipulator’s default joint impedance settings, with stiffness and damping gains for the six arm joints and the gripper set to $k_p = [512, 768, 768, 512, 384, 256, 512]$ and $k_d = [25.6, 25.6, 25.6, 25.6, 25.6, 25.6, 25.6]$, respectively. For the leg controller, we apply uniform PD gains of $k_p = 360$ and $k_d = 5$ across all twelve leg joints. The actor policy network is an MLP with hidden dimensions of [512, 256, 128], using the ELU activation function. The PPO hyperparameters, such as learning rate, are listed in \cite{rudin2022learning}.

\subsubsection{Hardware Platform}
An overview of the legged manipulator system is shown in Fig. \ref{fig:11}. During the real-world experiment, we use the Unitree B2 quadruped robot, equipped with a Unitree Z1 manipulator on the back, as our legged manipulator platform. The Z1 manipulator has a maximum payload of 2\,kg, which is far below the mass of the objects we need to move. The platform is outfitted with a Livox Mid360 LiDAR mounted on the rear of the base and a RealSense D455 RGB-D camera mounted on the robot’s head. All planning, control, and perception modules—including the RL policy, TAMP, localization, and object 6D pose detection—are executed on an onboard Intel Core i7 processor.

\subsubsection{Perception}
The perception module mainly consists of three parts: localization, object detection, and the object's 6D pose estimation. We utilize the lidar-based method \cite{ou2025pl} for online relocalization. For object detection, we manually collect data of the objects and train the model using YOLOv8 \cite{varghese2024yolov8}. For 6D pose estimation, we adopt the FoundationPose framework \cite{wen2024foundationpose} and deploy it to our own set of objects. During object manipulation, the relative position and orientation between the objects and the robot can be obtained through AprilTag \cite{olson2011apriltag} attached to the object.

\subsubsection{Object Grasping}
Based on the object pose estimated by the perception module and the predefined graspable regions of the object, the target grasp pose is computed via homogeneous transformations. The corresponding arm joint angles are then generated through inverse kinematics using Pinocchio \cite{carpentier2019pinocchio}. The real-time end-effector pose is subsequently sent to the low-level WBC, which coordinates the arm and leg motions to achieve stable whole-body object grasping.

\subsection{Comparison with Baselines}\label{subsec:Comparison}
To validate our approach, we compare the algorithm's performance with the following baselines:
\begin{itemize}

\item \textbf{Direct Method:} We assume the robot-object system as a single rigid body and employ planar kinematics to map the desired object velocity to the corresponding velocity command of the robot base. This baseline reveals the limitations of the direct control method.

\item \textbf{Vanilla Method:} This baseline employs only a simple MLP network with proprioceptive observations, without any additional velocity estimation or ICR module, as well as the object states. It is used to verify the necessity of incorporating object observations and other proposed modules.

\item \textbf{Physical Parameter Estimation (PPE) \cite{yang2025efficient}:} A related method that estimates physical parameters of objects (e.g., friction coefficient, mass, and center of mass) to enable the network to better model implicit object dynamics, resulting in more accurate and reliable object control.

\item \textbf{RobotMover \cite{li2025robotmover}:} This is the most similar prior work to ours, as it also incorporates object observations to train an object velocity controller. The comparison highlights the advantage of our approach, which can use a single policy to handle various objects.

\item \textbf{W/o Pre-Trained Controller:} To validate the effectiveness of our hierarchical training framework over a purely end-to-end design, we remove the low-level controller and directly train an object controller from scratch using the same observations as our proposed method.  

\item \textbf{W/o Velocity Estimator:} To examine the contribution of the proposed velocity estimator, we exclude its outputs from the observation space during training, while keeping all other settings unchanged.  

\item \textbf{W/o ICR Module:} To evaluate the importance of the proposed ICR module, we remove it from the architecture while keeping other components intact.  

\end{itemize}

 For a fair comparison, all baseline methods are trained in the same RL environment for all subsequent qualitative and quantitative evaluations.

\begin{figure}[t]
    \centering
    \includegraphics[width=0.95\linewidth]{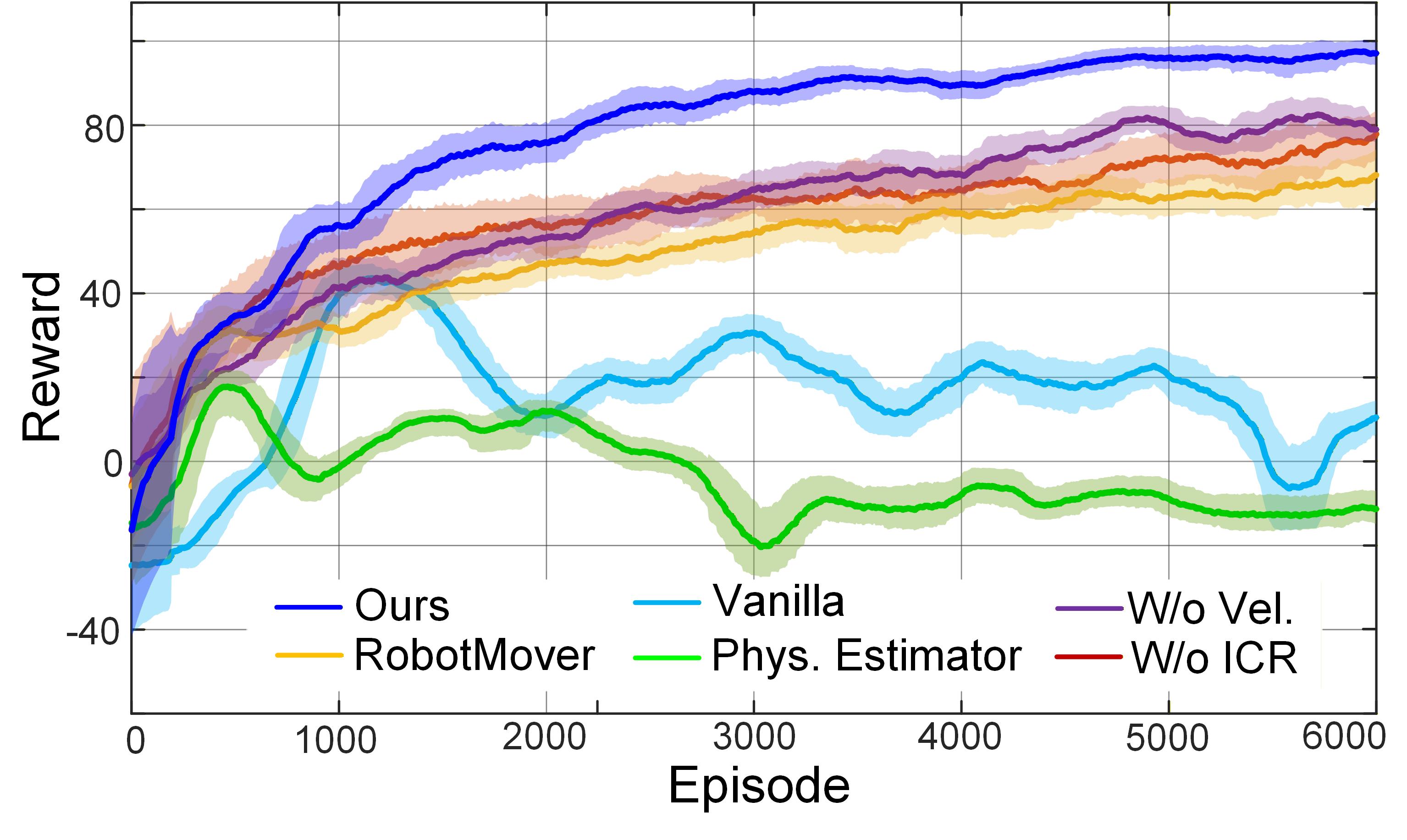}
    \caption{Comparison of training rewards over 6000 episodes against three baselines and two ablation variants.}
    \label{fig:reward}
\end{figure}

\begin{table*}[htbp]
  \centering
  \caption{
  % Comparison of Velocity Tracking Accuracy Measured by MAE and SD Across Various Kinds of Objects and Methods. Lower MAE and SD Indicate Better Tracking Performance.
  Quantitative Evaluation of Object Velocity Tracking Performance. The Table Reports the MAE and SD between the Commanded and Actual Object Velocities across Three Object Categories.
  }
  \begin{threeparttable}
  \resizebox{\textwidth}{!}{
    \begin{tabular}{cccccccccccc}
    \toprule
    \multirow{2}[4]{*}{\textbf{Method}} & 
    \multirow{2}[4]{*}{\textbf{Metric}} & 
    \multicolumn{3}{c}{\textbf{Bin}} & 
    \multicolumn{3}{c}{\textbf{Chair}} & 
    \multicolumn{3}{c}{\textbf{Table}} &
    \multirow{2}[4]{*}{\makecell{\textbf{Mean} \\ \textbf{Error}}} \\
    \cmidrule(lr){3-5} \cmidrule(lr){6-8} \cmidrule(lr){9-11} 
    % \multirow{2}[4]{*}{\textbf{Method}} & \multicolumn{3}{c}{\textbf{bin}} & \multicolumn{3}{c}{\textbf{Chair}} & \multicolumn{3}{c}{\textbf{Table}} \\
         & & $v_x$ (\text{m/s}) & $v_y$ (\text{m/s}) & $\omega_z$ (\text{rad/s}) & $v_x$ (\text{m/s}) & $v_y$ (\text{m/s}) & $\omega_z$ (\text{rad/s}) & $v_x$ (\text{m/s}) & $v_y$ (\text{m/s}) & $\omega_z$ (\text{rad/s}) \\
    \midrule

     Direct Method & \makecell{MAE \\ SD} & \makecell{0.194 \\ 0.0608} & \makecell{0.146 \\ 0.0452} & \makecell{0.0742 \\ \textcolor{red}{0.0605}} & \makecell{0.177 \\ 0.0357} & \makecell{0.126 \\ 0.0230} & \makecell{0.0823 \\ \textcolor{red}{0.0415}} & \makecell{0.184 \\ 0.0922} & \makecell{0.0933 \\ 0.0463} & \makecell{0.0889 \\ 0.0554} & \makecell{0.130 \\ 0.0512} \\

 Vanilla Method & \makecell{MAE \\ SD} & \makecell{0.204 \\ 0.0837} & \makecell{0.0912 \\ 0.0774} & \makecell{0.129 \\ 0.115} & \makecell{0.160 \\ 0.0499} & \makecell{0.0772 \\ 0.0220} & \makecell{0.0704 \\ 0.0328} & \makecell{0.211 \\ 0.0952} & \makecell{\textcolor{orange}{0.0393} \\ 0.0381} & \makecell{0.141 \\ 0.0761} & \makecell{0.124 \\ 0.0656} \\

   PPE \cite{yang2025efficient}& \makecell{MAE \\ SD} & \makecell{0.365 \\ 0.144} & \makecell{0.136 \\ 0.0893} & \makecell{0.163 \\ 0.181} & \makecell{0.0951 \\ 0.0475} & \makecell{0.0870 \\ 0.0356} & \makecell{0.202 \\ 0.0593} & \makecell{0.387 \\ 0.0565} & \makecell{0.111 \\ 0.0301} & \makecell{0.379 \\ 0.0603} & \makecell{0.214 \\ 0.0782} \\

    RobotMover \cite{li2025robotmover}& \makecell{MAE \\ SD} & \makecell{0.120 \\ 0.0797} & \makecell{0.0820 \\ 0.0500} & \makecell{0.108 \\ 0.103} & \makecell{0.0433 \\ \textcolor{orange}{0.0284}} & \makecell{\textcolor{orange}{0.0397} \\ \textcolor{orange}{0.0199}} & \makecell{\textcolor{orange}{0.0611} \\ 0.0511} & \makecell{0.0556 \\ 0.0468} & \makecell{0.0472 \\ 0.0329} & \makecell{0.0959 \\ 0.0686} & \makecell{0.0725 \\ 0.0534} \\

    % W/o Pre-trained Controller & 0.270 ± 0.161 & 0.0425 ± 0.0668 & 0.198 ± 0.199 & 0.269 ± 0.141 & 0.0755 ± 0.103 & 0.170 ± 0.177 & 0.273 ± 0.164 & 0.0272 ± 0.0475 & 0.164 ± 0.0961 & 0.176 ± 0.121 \\

    W/o Pre-Trained Controller & \textcolor{blue}{\ding{55}} & \textcolor{blue}{\ding{55}} & \textcolor{blue}{\ding{55}}  & \textcolor{blue}{\ding{55}}  & \textcolor{blue}{\ding{55}} & \textcolor{blue}{\ding{55}}  & \textcolor{blue}{\ding{55}}  & \textcolor{blue}{\ding{55}} & \textcolor{blue}{\ding{55}} & \textcolor{blue}{\ding{55}} & \textcolor{blue}{\ding{55}} \\

    W/o Vel. Estimator& \makecell{MAE \\ SD} & \makecell{\textcolor{orange}{0.0593} \\ 0.0497} & \makecell{\textcolor{orange}{0.0558} \\ 0.0334} & \makecell{\textcolor{red}{0.0713} \\ 0.0929} & \makecell{0.0596 \\ 0.0376} & \makecell{0.0619 \\ 0.0272} & \makecell{0.0892 \\ 0.0465} & \makecell{0.0463 \\ 0.0442} & \makecell{\textcolor{red}{0.0382} \\ \textcolor{orange}{0.0228}} & \makecell{\textcolor{orange}{0.0568} \\ 0.0499} & \makecell{0.0598 \\ \textcolor{orange}{0.0449}} \\

    W/o ICR Module& \makecell{MAE \\ SD} & \makecell{0.0660 \\ \textcolor{orange}{0.0491}} & \makecell{0.0606 \\ \textcolor{orange}{0.0299}} & \makecell{0.0849 \\ 0.0995} & \makecell{\textcolor{orange}{0.0424} \\ 0.0325} & \makecell{0.0596 \\ 0.0277} & \makecell{0.0676 \\ 0.0507} & \makecell{\textcolor{orange}{0.0444} \\ \textcolor{orange}{0.0425}} & \makecell{0.0489 \\ 0.0245} & \makecell{0.0625 \\ \textcolor{orange}{0.0497}} & \makecell{\textcolor{orange}{0.0597} \\ 0.0451} \\

    % \textbf{Ours} & \textbf{0.0456} ± \textbf{0.0349} & \textbf{0.0507} ± \textbf{0.0197 }& 0.0727 ± \textbf{0.0718}& \textbf{0.0370} ± \textbf{0.0277 }& 0.0410 ± \textbf{0.0169}& 0.0726 ± \textbf{0.0446}& 0.0478 ± \textbf{0.0289} & \textbf{0.0381} ± \textbf{0.0181} & 0.0594 ± \textbf{0.0422} & \textbf{0.0522} ± \textbf{0.0302} \\

    \textbf{Ours}& \makecell{MAE \\ SD} & \makecell{\textcolor{red}{0.0456} \\ \textcolor{red}{0.0349}} & \makecell{\textcolor{red}{0.0507} \\ \textcolor{red}{0.0197}} & \makecell{\textcolor{orange}{0.0727} \\ \textcolor{orange}{0.0718}} & \makecell{\textcolor{red}{0.0409} \\ \textcolor{red}{0.0236}} & \makecell{\textcolor{red}{0.0343} \\ \textcolor{red}{0.0157}} & \makecell{\textcolor{red}{0.0585} \\ \textcolor{orange}{0.0423}} & \makecell{\textcolor{red}{0.0323} \\ \textcolor{red}{0.0250}} & \makecell{0.0536 \\ \textcolor{red}{0.0195}} & \makecell{\textcolor{red}{0.0490} \\ \textcolor{red}{0.0344}} & \makecell{\textcolor{red}{0.0486}  \\ \textcolor{red}{0.0319}} \\
    
    % \midrule
    % \textbf{Improve} & 23.1\% / 28.9\% &  9.14\% /  34.1\%   & -1.96\% / 22.7\% & 12.7\% / 14.8\% & -1.99\% / 19.9\% & -18.0\% / 4.09\% & -3.24\% / 24.9\% & 0.26\% / 20.6\% & -4.58\% / 15.1\% & \textbf{13.6}\% / \textbf{25.1}\%\\
    \bottomrule
    \end{tabular}
  }
    \begin{tablenotes}
    \footnotesize
    \item \text{Note:} \textcolor{blue}{\ding{55}} indicates that the method fails to complete the task. \textcolor{red}{Red} indicates the best results, \textcolor{orange}{orange} indicates the second-best results.
    \end{tablenotes}
    \end{threeparttable}
    \label{tab:velocity_tracking}
\end{table*}

\begin{figure*}[t]	
	\centering
\includegraphics[width=1.0\linewidth]{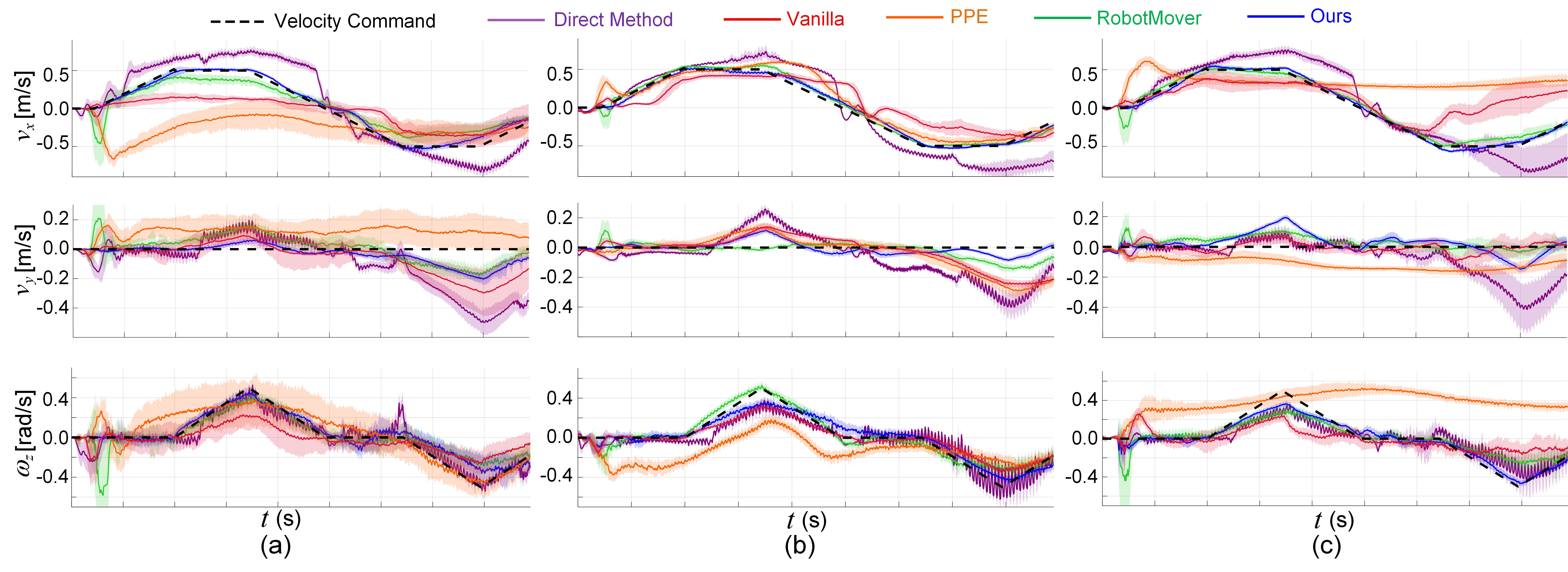}
	\setlength{\abovecaptionskip}{-0pt} 
	\caption{Comparison of velocity tracking curves of bin (a), chair (b), and table (c) under five kinds of methods.
    }
	\label{fig:4}
    % \vspace{-8pt}
\end{figure*}

\subsubsection{Comparison of Learning Performance}
Fig. \ref{fig:reward} shows the RL reward curves of six different object controllers over 6000
episodes. Among them, the proposed approach achieves the highest average reward. The second-best performers are the two ablation methods, followed by the RobotMover method. The worst-performing algorithms are the vanilla approach and the PPE. Compared to the two ablation methods, our approach incorporates both the ICR module and the object velocity prediction module. The results show that both modules contribute to varying degrees of improvement in the reward, demonstrating that both the ICR module and the object velocity estimator are beneficial in training the object velocity controller. In contrast, the RobotMover method lacks both of these modules, leading to a lower average reward. In addition, both the vanilla method and the physical parameter prediction-based method fail to converge. The vanilla method, in particular, relies solely on the robot's proprioceptive inputs without any object-specific information, which may hinder its ability to learn how to accurately control different objects. This limitation, coupled with the simple network structure, makes it difficult to handle a diverse range of objects. The PPE aims to enhance the controller's performance by predicting physical parameters of objects, such as mass and friction coefficient. Although it performs well in tasks involving a single object \cite{yang2025efficient}, it yields the poorest results in tasks involving a variety of objects. A likely explanation is that, compared with a single object, the training set spans a much wider range of physical properties, making accurate parameter estimation challenging. Such unreliable predictions, in turn, reduce the overall reward.

\subsubsection{Comparison of Object Controllers}
We evaluate our proposed high-level object velocity controller on multiple object types, including chairs, tables, and bins. For each object type, we randomly generate 30 environments as shown in Table \ref{tab:randomize}. Under the same desired velocity commands, we record the objects’ actual velocities and compute the mean absolute error (MAE) and standard deviation (SD) between the commanded and actual velocities as evaluation metrics across baselines. Table \ref{tab:velocity_tracking} and Fig. \ref{fig:4} show the results quantitatively and qualitatively. 

 \begin{figure*}[t]	
	\centering
    % \hspace*{0.0cm}
\includegraphics[width=0.95\linewidth]{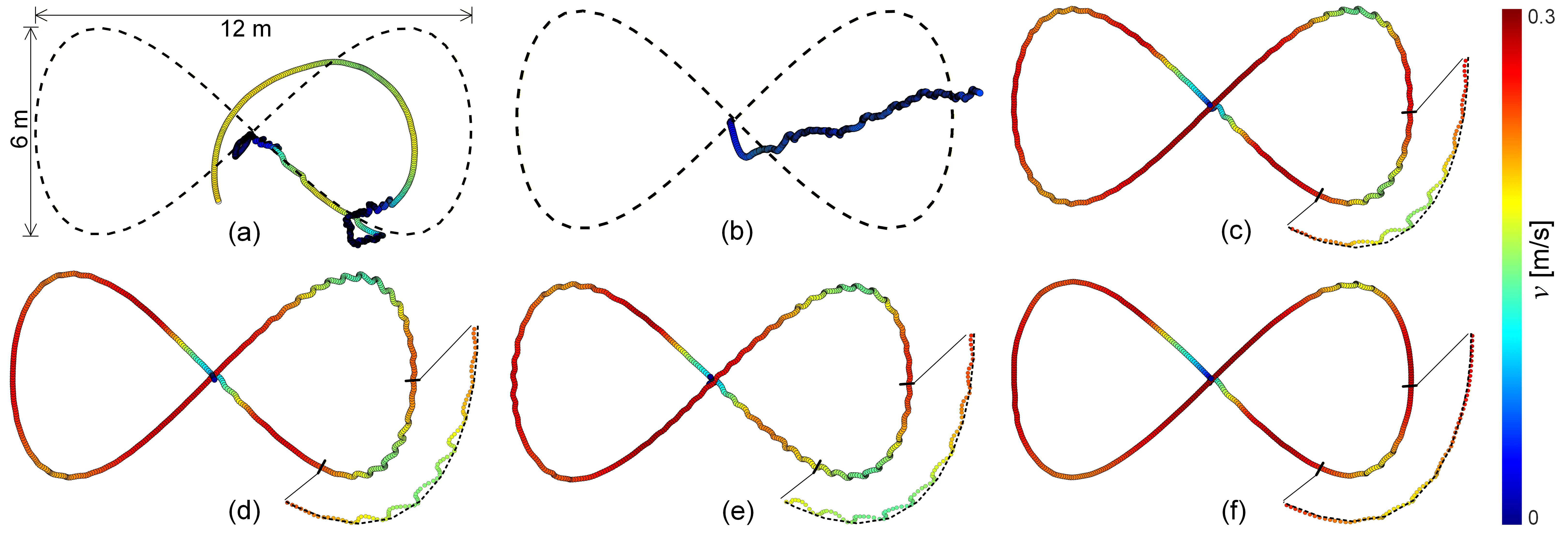}
	\setlength{\abovecaptionskip}{-0pt} 
	\caption{Comparison of reference tracking performance across different object velocity controllers: (a) Vanilla method, (b) PPE, (c) RobotMover, (d) without the Velocity Estimator, (e) without the ICR module, and (f) our method. The dashed black curves indicate the reference trajectories, while the object trajectories are color-coded according to the velocity.
    }
	\label{fig:traj_tracking}
    % \vspace{-8pt}
\end{figure*}

As illustrated in Table \ref{tab:velocity_tracking}, red entries indicate the lowest tracking error and orange entries indicate the second-lowest. Overall, our method achieves the best performance on most metrics across the three types of objects. For the direct method, we observe consistently large MAE values, indicating poor object velocity tracking. This is mainly because the assumption that the robot-object system behaves as a single rigid body during planar velocity mapping does not strictly hold in practice due to the joint torque limits of the manipulator. Besides, the desired object velocity is converted into a base velocity command for the low-level WBC, whose own velocity tracking is also imperfect. Moreover, since the direct method does not explicitly account for the object state, the object may also topple during pushing and pulling, a phenomenon rarely observed in other RL-based baselines, which further degrades tracking accuracy. For the vanilla method, we also observe relatively large tracking errors, mainly due to its limited proprioceptive inputs and the overly simple network architecture, which is consistent with the aforementioned analysis. 
Unlike the vanilla method, the RobotMover approach achieves lower tracking errors by including the object state as observation. However, its policies are object-specific and do not generalize well across multiple object types. As a result, the SD of velocity tracking is \textbf{67.4\%} higher than the proposed method. 
Consistent with the average reward curves, the PPE aims to enhance control accuracy by explicitly predicting the objects’ physical parameters \cite{yang2025efficient}. However, this method exhibits the largest tracking error among all baselines. While this approach is effective for a single object with fixed physical properties, it becomes unreliable when extended to multiple objects. A detailed analysis indicates that the main challenge stems from the diversity of object dynamics: objects with different physical properties correspond to distinct dynamic models, which makes accurate parameter prediction difficult. These inaccurate predictions, in turn, negatively affect the object velocity tracking performance.

\begin{table}[t]
  \centering
  \caption{Performance Comparison of Reference Tracking}
  \resizebox{0.95\columnwidth}{!}{%
    \begin{tabular}{cccc}
    \toprule
    \text{Methods}  & 
    \makecell{Max. Abs. \\  Error (m)} &
    \makecell{Mean Abs. \\  Error (m)} &
    \makecell{Completion \\ Time (s)} \\
    \midrule
    Vanilla Method  & \textcolor{blue}{\ding{55}} & \textcolor{blue}{\ding{55}} & \textcolor{blue}{\ding{55}}   \\
    PPE   & \textcolor{blue}{\ding{55}} & \textcolor{blue}{\ding{55}} & \textcolor{blue}{\ding{55}}  \\
    RobotMover   & 0.162 & 0.0422 & 283   \\
    W/o Vel. Estimator  & 0.176 & 0.0354 & 300   \\
    W/o ICR Module   & 0.148 & 0.0447 & 295  \\
   \textbf{Our Method}   & \textcolor{red}{0.0764} &  \textcolor{red}{0.0241} &  \textcolor{red}{267}   \\
    \bottomrule
    \end{tabular}%
  }
   \begin{tablenotes}
    \footnotesize
    \item \text{Note:} \textcolor{blue}{\ding{55}} indicates task failure, and \textcolor{red}{red} denotes the best result.
    \end{tablenotes}
  \label{tab:ref_tracking}
   \vspace{-3pt}
\end{table}

In addition to the baseline methods above, we also conduct several ablation studies to further examine the key components in our framework. If the low-level pre-trained controller is not utilized, training an object-level controller fails, since learning locomotion and object manipulation simultaneously is highly challenging and typically requires reward terms that are inherently conflicting. Additionally, removing the object-velocity estimator leads to a noticeable degradation in controller performance. Likewise, disabling the ICR module increases both the velocity tracking error and its SD. These results further confirm that both the velocity estimator and the ICR module play a crucial role in improving object velocity tracking accuracy. Overall, our approach outperforms all the baseline methods. Compared to the second-best method, the proposed method reduces MAE by \textbf{22.8\%} and SD by \textbf{40.8\%}.

To qualitatively illustrate the velocity tracking result, Fig. \ref{fig:4} shows the objects' velocity tracking curves of different baselines. Fig. \ref{fig:4}(a)-(c) illustrate the linear and angular velocity tracking curves for the bin, chair, and table, respectively. In this figure, the black dashed line denotes the object velocity command, while the purple, red, orange, green, and blue lines represent the ground-truth mean velocities of 30 objects under five different methods. The shaded region around the line corresponds to the SD. For these three kinds of objects, it can be qualitatively observed that our method achieves the highest velocity tracking accuracy and demonstrates the smallest SD compared to other baseline methods.

\begin{figure*}[t]	
	\centering
        % \vspace{3pt}
	\includegraphics[width=0.98\linewidth]{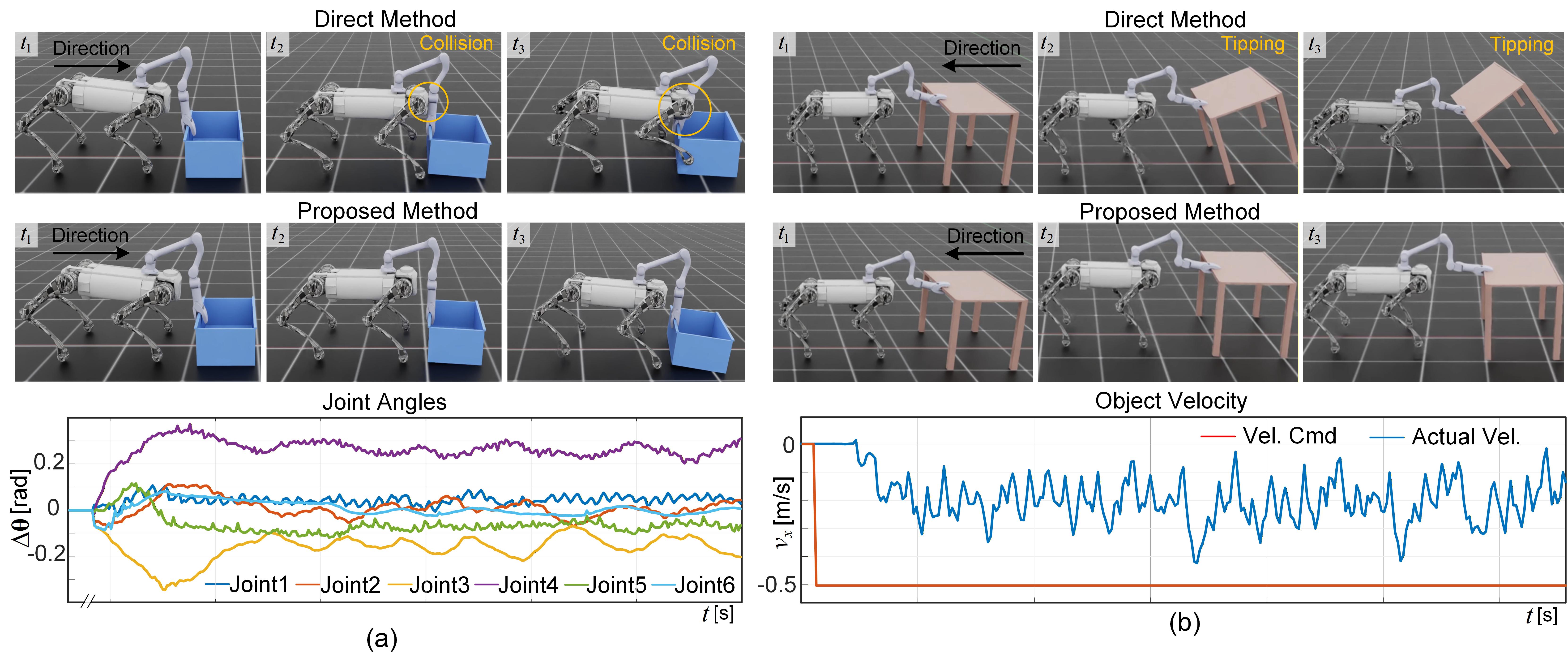}
	\setlength{\abovecaptionskip}{-5pt} 
	\caption{Comparison of object control between the proposed method and the direct method. (a) Safety assurance for self-collision prevention. (b) Adaptive stabilization against object toppling.
    }
	\label{fig:5}
\end{figure*}

\begin{figure}[t]	
	\centering
    % \vspace{-10pt}
	\includegraphics[width=0.95\linewidth]{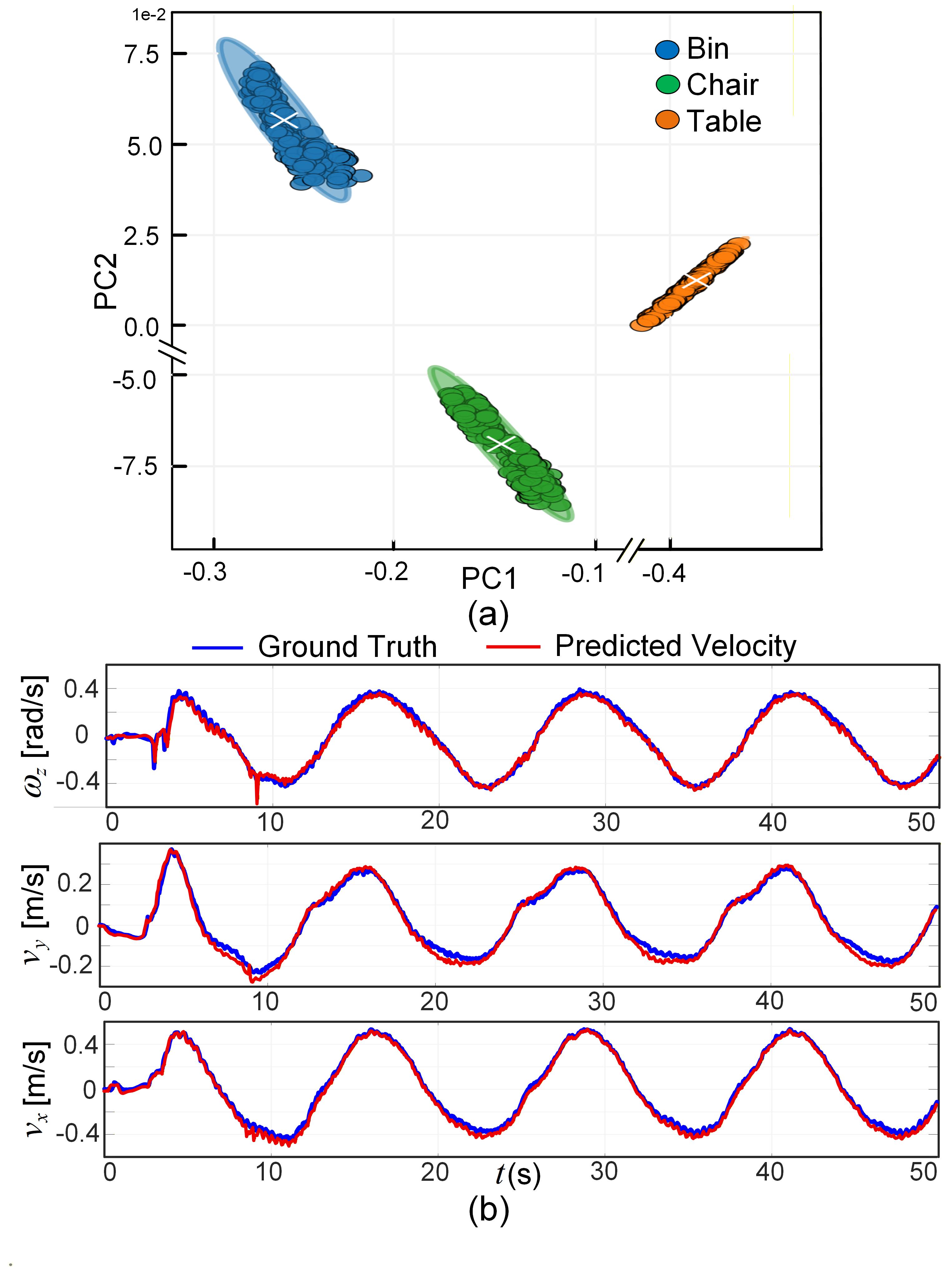}
	\setlength{\abovecaptionskip}{-15pt} 
	\caption{(a) PCA visualization of ICR module features with 95\% confidence ellipses across three object categories. (b) Comparison between predicted and ground truth velocities of the table.
    }
	\label{fig:6}
\end{figure}

\subsubsection{Comparison of the Reference Tracking Performance}
As illustrated in Fig. \ref{fig:traj_tracking}, we conduct reference tracking experiments using six different methods to evaluate our proposed object controller. The reference trajectory is a figure-eight path with a length of 12\,m and a width of 6\,m, which the robot is required to track at a constant speed of 0.3\,m/s. Under identical control parameter settings, we record the resulting object trajectories generated by each method.
The results show that, due to large open-loop velocity tracking errors, the object controllers trained using the vanilla method and the PPE method fail to accomplish the reference tracking task, even when operating in a closed-loop manner. In contrast, the controllers based on methods (c), (d), (e), and (f) are able to successfully complete the task. 
From a qualitative perspective, the proposed controller demonstrates improved reference tracking accuracy with consistently smaller deviations, as evidenced in the zoomed-in regions in Fig. \ref{fig:traj_tracking}. From a quantitative perspective, as reported in Table~\ref{tab:ref_tracking}, the proposed method delivers the best overall performance across all baselines, achieving the lowest maximum absolute tracking error, the lowest MAE, and the shortest completion time. Notably, the MAE is reduced by \textbf{85.5\%} relative to the second-best method.

These results highlight that, in our system, open-loop velocity tracking accuracy has a significant impact on closed-loop reference tracking performance, which is particularly critical when controlling objects through narrow passages.

\subsubsection{Comparison with the Direct Method}
As shown in Fig.~\ref{fig:5}, we conduct two experiments to highlight the advantage of our method over the direct method during large object manipulation. In the experiment, for both bins and tables, the object-ground dynamic friction coefficient is set to $0.6$, and the object mass is set to 15\,kg. The input to the system is the desired object velocity $v_x^b$: we set $v_x^b=0.5$\,m/s for the bin and $v_x^b=-0.5$\,m/s for the table. The robot then performs velocity control using either a direct velocity-mapping baseline (\textit{Direct Method}) or our proposed object controller. 
In both cases, the key failure mode of the direct method is the same: it rigidly enforces the commanded base velocity without accounting for interaction-induced constraints, leading to a growing mismatch between the robot motion and the object response. This mismatch drives the manipulator toward unsafe configurations, either exceeding torque limits and causing self-collision (Fig.~\ref{fig:5}(a)) or inducing object toppling when pulling (Fig.~\ref{fig:5}(b)) due to ignoring object pose feedback. 
In contrast, our proposed object controller is trained with self-collision avoidance constraint (Eqs.~(\ref{eq:collision})), hardward-safe actuation constraint (Eqs.~(\ref{eq:effort})), as well as the object pose stability penalties (Table~\ref{tab:reward}). These designs enable the robot to adjust arm joint angles to avoid self-collision (Fig.~\ref{fig:5}(a)) and, when necessary, to trade off object-velocity tracking for safety to stabilize the object against toppling (Fig.~\ref{fig:5}(b)). Videos are provided in \textcolor{red}{Supplementary Video S1}.

\subsection{Visualization of ICR Module and Object Velocity Estimator}\label{subsec:ICR}
Beyond the comparisons with baseline methods, we further validate the effectiveness of the proposed two key modules through qualitative visualizations.

\begin{table}[t]
  \centering
  \caption{Error Analysis of the Object Velocity Estimator}
  \resizebox{1.0\columnwidth}{!}{
    \begin{tabular}{ccccccc}
    \toprule
    \multirow{2}[4]{*}{\textbf{Objects}} & 
    \multicolumn{2}{c}{$v_x$ (m/s)} & 
    \multicolumn{2}{c}{$v_y$ (m/s)} & 

    \multicolumn{2}{c}{$\omega_z$ (rad/s)} \\
    \cmidrule(lr){2-3} \cmidrule(lr){4-5} \cmidrule(lr){6-7}
         & RMSE & MAE
         & RMSE & MAE
         & RMSE & MAE \\
    \midrule
    Bin 
        & 0.0253 & 0.0189
        & 0.0193 & 0.0161
        & \textcolor{blue}{0.0650} & \textcolor{blue}{0.0535} \\
        
    Chair
        & 0.0215 & 0.0177
        & \textcolor{red}{0.0101} & \textcolor{red}{0.0081}
        & 0.0166 & 0.0132 \\
        
    Table
        & 0.0188 & 0.0144
        & 0.0179 & 0.0146
        & 0.0237 & 0.0184\\
    \bottomrule
    \end{tabular}
}
    \begin{tablenotes}
    \footnotesize
      \item \text{Note:} \textcolor{red}{Red} denotes the best results, while \textcolor{blue}{blue} denotes the worst.
    \end{tablenotes}
  \label{tab:table2}
\end{table}

\subsubsection{Visualization of ICR Module}
To evaluate whether the ICR module can distinguish interactions between the legged manipulator and different objects, we employ Principal Component Analysis (PCA) to compress the dimensionality of the graph-level feature $\mathbf{l}_t$ extracted from the GNN and visualize the resulting embeddings. As illustrated in Fig. \ref{fig:6}(a), a total of 900 interaction scenarios are tested, with 300 for each object category. The feature $\mathbf{l}_t \in \mathbb{R}^{128}$ is projected onto the first two principal components, PC1 and PC2. In the figure, blue, green, and orange circles represent the projected features of bins, chairs, and tables, respectively. The result reveals tight and well-separated clusters for each object category in the PCA space, indicating high intra-class consistency and large inter-class divergence. This demonstrates that the ICR module can effectively distinguish different interaction configurations.
Within the same category, the features may not fully overlap, likely due to randomness in training conditions (e.g., friction coefficients and object masses).

\begin{figure}[t]	
	\centering
    % \vspace{-10pt}
	\includegraphics[width=1\linewidth]{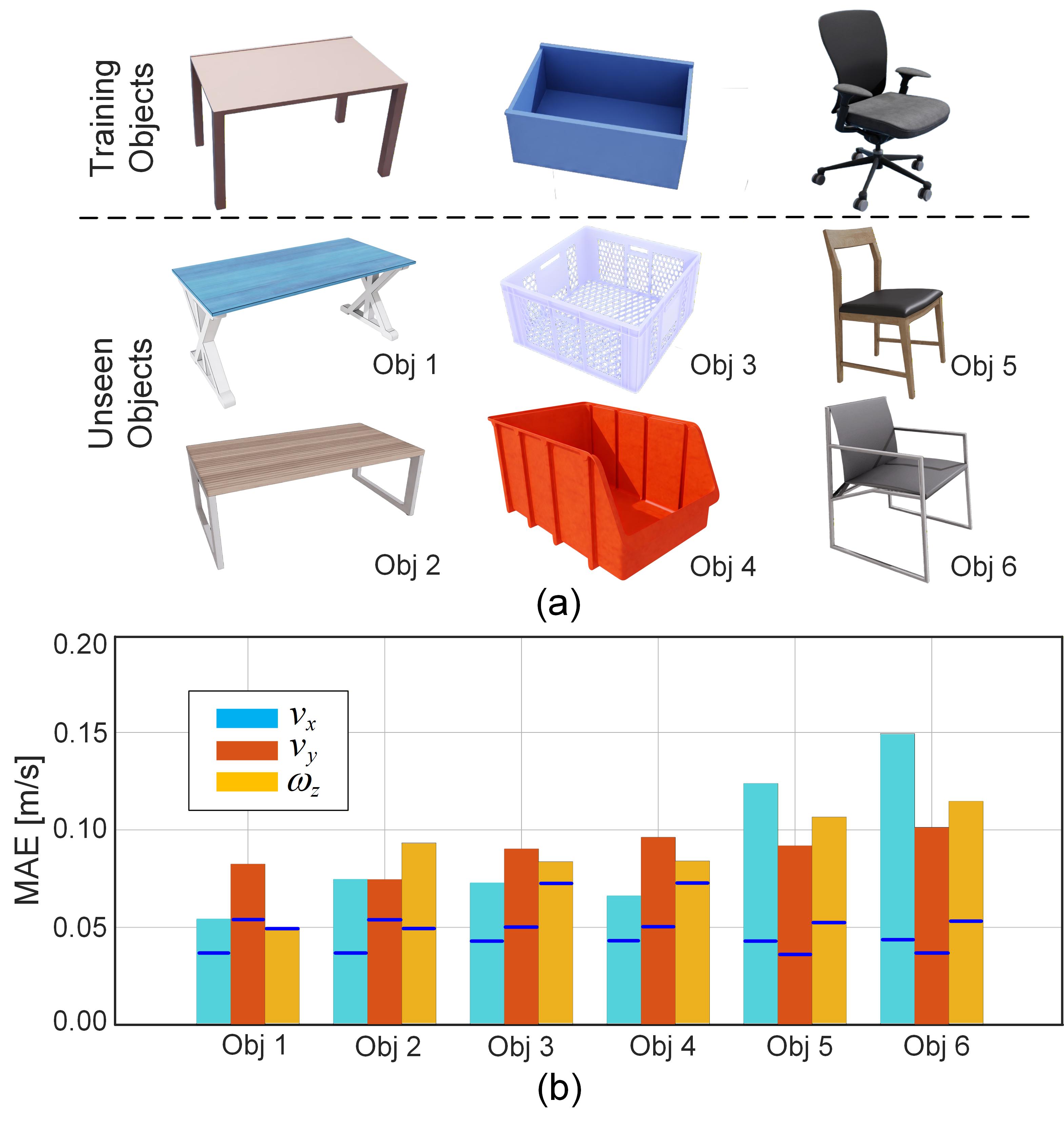}
	\setlength{\abovecaptionskip}{-15pt} 
	\caption{Generalization performance of the trained object controller on six previously unseen objects in velocity-tracking experiments. (a) shows three training objects and six unseen objects. (b) The bars in three different colors denote the MAE of velocity tracking for six unseen objects, and the blue line segment indicates the best result achieved on seen objects of the same type. 
    }
	\label{fig:multiObj}
\end{figure}

\begin{figure*}[t]	
	\centering
        % \vspace{3pt}
	\includegraphics[width=0.95\linewidth]{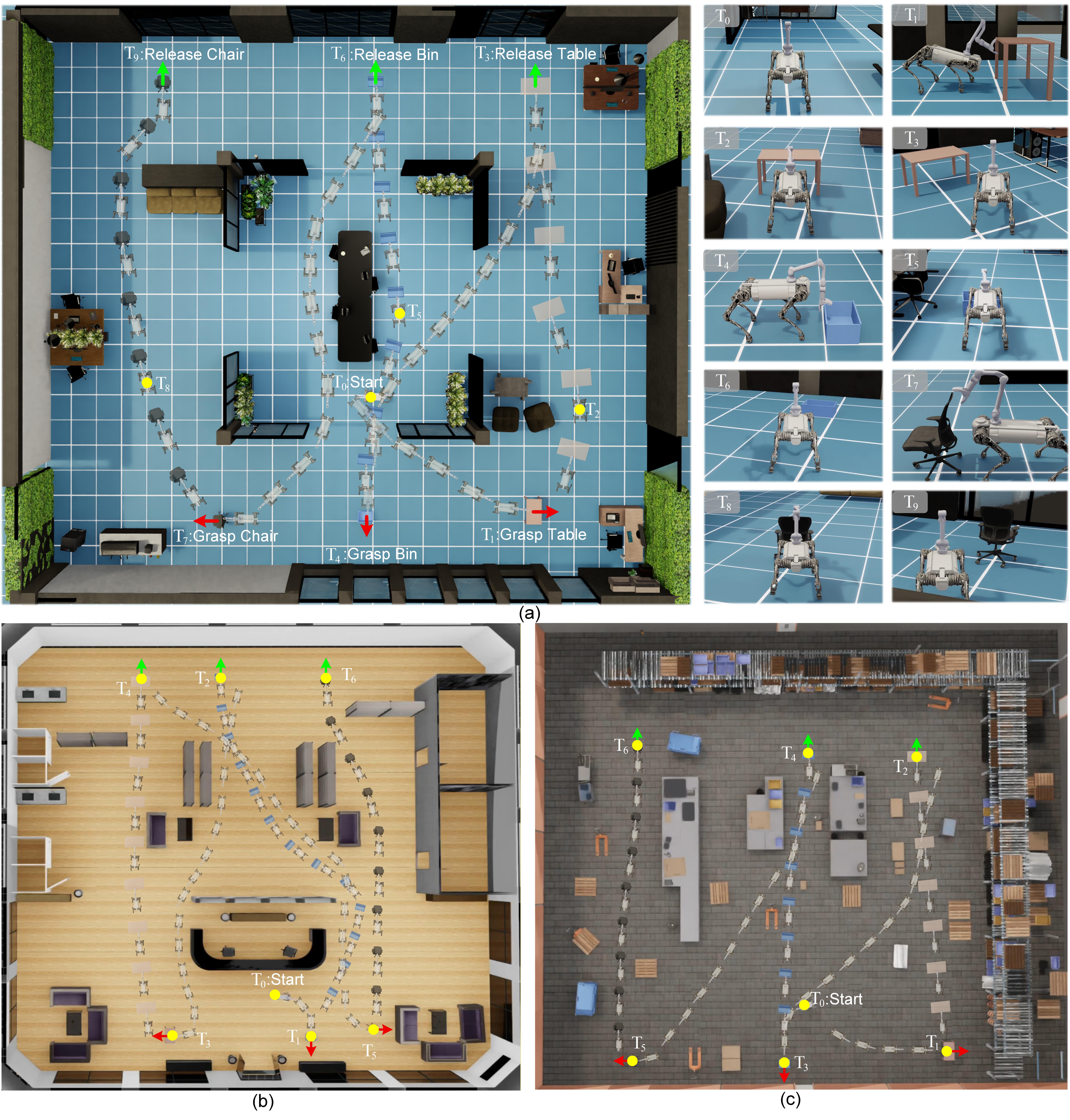}
	\setlength{\abovecaptionskip}{-0pt} 
	\caption{Simulation results of various object rearrangements in three large-scale scenarios: (a) office, (b) library, and (c) warehouse. Take office as an example, 
    starting at $\text{T}_0$, the legged manipulator grasps and releases the table at $\text{T}_1$ and $\text{T}_3$, 
    the bin at $\text{T}_4$ and $\text{T}_6$, and the chair at $\text{T}_7$ and $\text{T}_9$. 
    The insets on the right illustrate the feature motions across key timestamps ($\text{T}_0$-$\text{T}_9$). }
	\label{fig:7}
    \vspace{-6pt}
\end{figure*}

\subsubsection{Visualization of Velocity Estimator}
To evaluate the object velocity estimator, we applied sinusoid-like control signals to three categories of objects and recorded their ground-truth velocities along with the predicted velocities. As shown in Fig. \ref{fig:6}(b), the blue line represents the ground truth velocity, the red line denotes the predicted velocity. Overall, although small fluctuations in prediction may occur when the velocity changes sharply, the predicted curves closely overlap with the ground-truth curves. Meanwhile, for quantitative analysis, we compute both Root Mean Square Error (RMSE) and MAE between predicted and actual velocities for all three object categories. As summarized in Table \ref{tab:table2}, all error metrics remain at the order of $10^{-2}$, which further confirms that the trained velocity estimator can accurately estimate various object velocities.

\subsection{Generalize to Unseen Objects}\label{subsec:unseenObj}
To evaluate the generalization capability of the learned policy, we conduct velocity-tracking experiments on six previously unseen objects, as shown in Fig. \ref{fig:multiObj}. The experiment setting is similar to that in Sec. \ref{subsec:Comparison}(2). For each object type, we randomly generate 30 environments and, under the same desired velocity commands, record the object’s actual velocity and compute the MAE.
We find that, compared with the best performance on seen objects of the same type, the MAE for Objects 1-4 increases by about 0.025\,m/s on average. Despite the substantial geometric differences between these unseen objects and the training objects, the overall velocity-tracking MAE still remains below 0.1\,m/s, with the average error along the $v_x$ direction below 0.075\,m/s. This level of control accuracy is comparable to that of \textit{RobotMover} baseline \cite{li2025robotmover} on seen objects and is sufficient for reliable path tracking (Fig. \ref{fig:traj_tracking}(c)), which highlights the robustness and generalization of our proposed method.

However, for objects 5 and 6, the velocity-tracking MAE is relatively larger, which increases by nearly 0.06\,m/s on average compared with the best performance on seen objects of the same type. A plausible explanation is that the training chair is equipped with passive wheels, resulting in predominantly rolling friction at the contact interface, whereas objects 5 and 6 interact with the ground through sliding friction. This mismatch leads to significantly different interaction dynamics, thereby degrading the tracking performance.

Overall, these results suggest that the proposed policy exhibits meaningful generalization to previously unseen objects with similar interaction configurations, and that its generalization performance is better when the object-ground friction models are consistent with those encountered during training.

\begin{table}[t]
  \centering
  \caption{Comparison Between the Greedy Planning Baseline and Our Proposed Task Planning Method.}
  \resizebox{0.85\columnwidth}{!}{%
    \begin{tabular}{ccccc}
    \toprule
    \textbf{Methods} & \textbf{Metric} & \textbf{Bin} & \textbf{Chair} & \textbf{Table} \\
    \midrule
    \multirow{2}{*}{TP (\textbf{Ours})} 
        & Time (s) & \textbf{224} & \textbf{154} & \textbf{191} \\
        & Distance (m)    & \textbf{86.51} & \textbf{59.53} & \textbf{81.88} \\
    \cmidrule(lr){1-5}
    
    \multirow{2}{*}{Greedy} 
        & Time (s) & 232 & 166 & 221 \\
        & Distance (m)  & 98.38 & 69.40 & 102.99 \\
    \bottomrule
    \end{tabular}%
  }
  \label{tab:task_planning_ablation}
\end{table}

\subsection{Evaluation of the Proposed Task Planning}\label{sec:eva_task_planning}
To further validate the effectiveness of the proposed task planning method, we conducted experiments in three settings: rearranging four chairs, four tables, and four bins in the office scenario \textcolor{red}{(Supplementary Video S2)}. In the experiments, the objects’ initial and target poses are kept fixed. Starting from the same initial configuration, the robot transports the four objects sequentially in a single continuous run under two different strategies, namely the greedy baseline and our proposed task planning method, and moves each object to its corresponding target pose.  We compare the task completion time and the total path length between the greedy strategy and the proposed task planning strategy, and summarize the results in Table \ref{tab:task_planning_ablation}. The results show that our proposed method achieves lower execution time and shorter total distance. This is because the greedy baseline always selects the nearest next object, while our method seeks a globally optimal rearrangement sequence.

\subsection{Various Object Rearrangement in Diverse Scenarios}\label{subsec:simexp}
As illustrated in Fig. \ref{fig:7}, it demonstrates the whole process of arranging the table, chair, and bin to their target poses in three large-scale scenarios. The red and green arrows denote the initial and goal poses of the objects, respectively. In this case, we first define the initial poses of the objects and the robot, and the objects' goal poses. Then, the proposed task planning module is utilized to generate the task sequence of object rearrangement. Afterward, the robot starts to grasp and move objects to their target poses one by one, as the process in Fig. \ref{fig:1}(a). Take the office scenario as an example, to clearly present the results, we extract ten key frames from $\text{T}_0$ to $\text{T}_9$. Specifically, the right inset at $\text{T}_0$ shows the third-person view of the robot at its initial position. The insets of $\text{T}_1, \text{T}_4$, and $\text{T}_7$ illustrate how the robot grasps objects using the low-level WBC, with $\text{T}_1$ highlighting the detailed grasping process. Furthermore, the insets of $\text{T}_3, \text{T}_6$, and $\text{T}_9$ depict the object release from a third-person perspective. Besides, the insets of $\text{T}_2, \text{T}_5$, and $\text{T}_8$ provide complementary third-person views of the robot manipulating three types of objects. It is worth noting that this task is completed under a single control policy, further demonstrating the generalization capability of the proposed high-level object controller. The video can be found in the \textcolor{red}{Supplementary Video S3.}

In addition, to further evaluate the robustness of the proposed framework, we assess the success rates of various object rearrangement tasks across these three different scenarios (\textcolor{red}{Supplementary Video S3}). The results are summarized in Table~\ref{tab:succ_rate}. A trial is successful if the task is completed without collisions; success rates are computed over 10 trials. Overall, the proposed framework is robust, achieving an average success rate of over 93\%. However, occasional table failures arise mainly from limited end-effector tracking accuracy in the low-level WBC during grasping.

\begin{table}[t]
  \centering
  \caption{Success Rates of Various Object Rearrangement across Three Simulated Scenarios}
  \resizebox{0.95\columnwidth}{!}{%
    \begin{tabular}{ccccc}
    \toprule
    \textbf{Scenarios}  & \textbf{Bin} & \textbf{Chair} & \textbf{Table} & \textbf{Average}\\
    \midrule
    Office (20\,m $\times$ 20\,m) & 100\% & 100\% & 90\%  & 96.7\% \\
    Library (30\,m $\times$ 30\,m)  & 100\% & 100\% & 80\% & 93.3\% \\
    Warehouse (40\,m $\times$ 40\,m)  & 100\% & 100\% & 90\%  & 96.7\% \\
    \bottomrule
    \end{tabular}%
  }
  \label{tab:succ_rate}
\end{table}

\subsection{Real-World Experiments}\label{subsec:realexp}
We conduct extensive real-world experiments to evaluate the proposed large-object rearrangement system. The experiments include rearrangement of diverse objects, long-horizon chair rearrangement, an autonomous long-distance object rearrangement task, and tests under varying ground conditions and payloads. We employ motion capture only to provide object pose estimates in the first two experiments; all other modules across all experiments rely solely on the robot’s onboard sensing. These experiments aim to: (1) demonstrate the ability of a single policy to generalize across different objects; (2) highlight the strong robustness of our entire system; (3) validate our system’s adaptability to diverse scenarios.

\subsubsection{Various Objects Rearrangement}
\begin{figure}[t]	
	\centering
    % \vspace{-10pt}
	\includegraphics[width=1.0\linewidth]{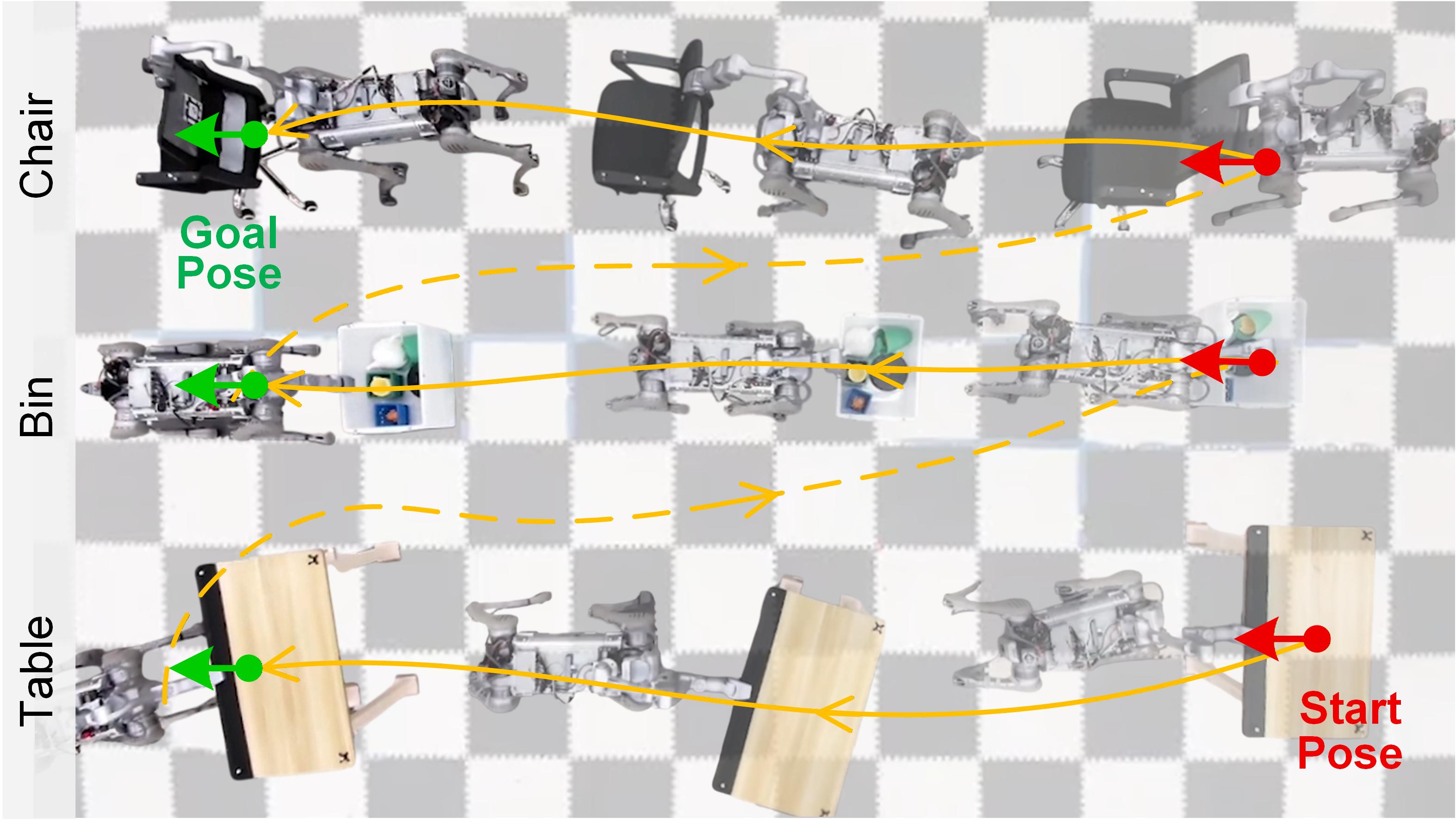}
	\setlength{\abovecaptionskip}{-0pt} 
	\caption{
    % Continuous rearrangement of different objects using a single object controller in an $8\,\text{m} \times 5\,\text{m}$ environment.
    Real-world demonstration of continuous object rearrangement. Our single object controller autonomously manages multiple objects in a $8\,\text{m} \times 5\,\text{m}$ physical environment.}
	\label{fig:diff_obj}
    \vspace{-5pt}
\end{figure}

As shown in Fig. \ref{fig:diff_obj}, it presents the experimental results of the robot sequentially rearranging a table (12\,kg), bin with an extra payload (6\,kg), and chair (8\,kg) to their target poses \textcolor{red}{(Supplementary Video S4)}. In the figure, the red arrows indicate the initial poses of the three objects, and the green arrows represent the target poses. Solid orange lines show the trajectories of the object centroids, dashed orange lines the trajectory of the robot centroid, and orange arrows the direction of motion. It is noteworthy that three different objects are controlled with only a single controller after grasping. In addition, one of the advantages of our system is that it supports bidirectional manipulation by design, since training explicitly covers both forward pushing and backward pulling. This allows the same controller to adapt its interaction strategy to the object’s contact condition, rather than relying on a fixed manipulation mode. In this experiment, the bin and the table experience high friction against the floor, which makes forward pushing difficult to regulate and often ineffective for achieving precise velocity tracking. Our system can switch to backward pulling to follow the planned trajectories in a stable manner. By contrast, for objects with lower effective friction, such as wheeled office chairs, the system can push efficiently to accomplish rearrangement.

\begin{figure*}[t]	
	\centering
        % \vspace{3pt}
	\includegraphics[width=0.98\linewidth]{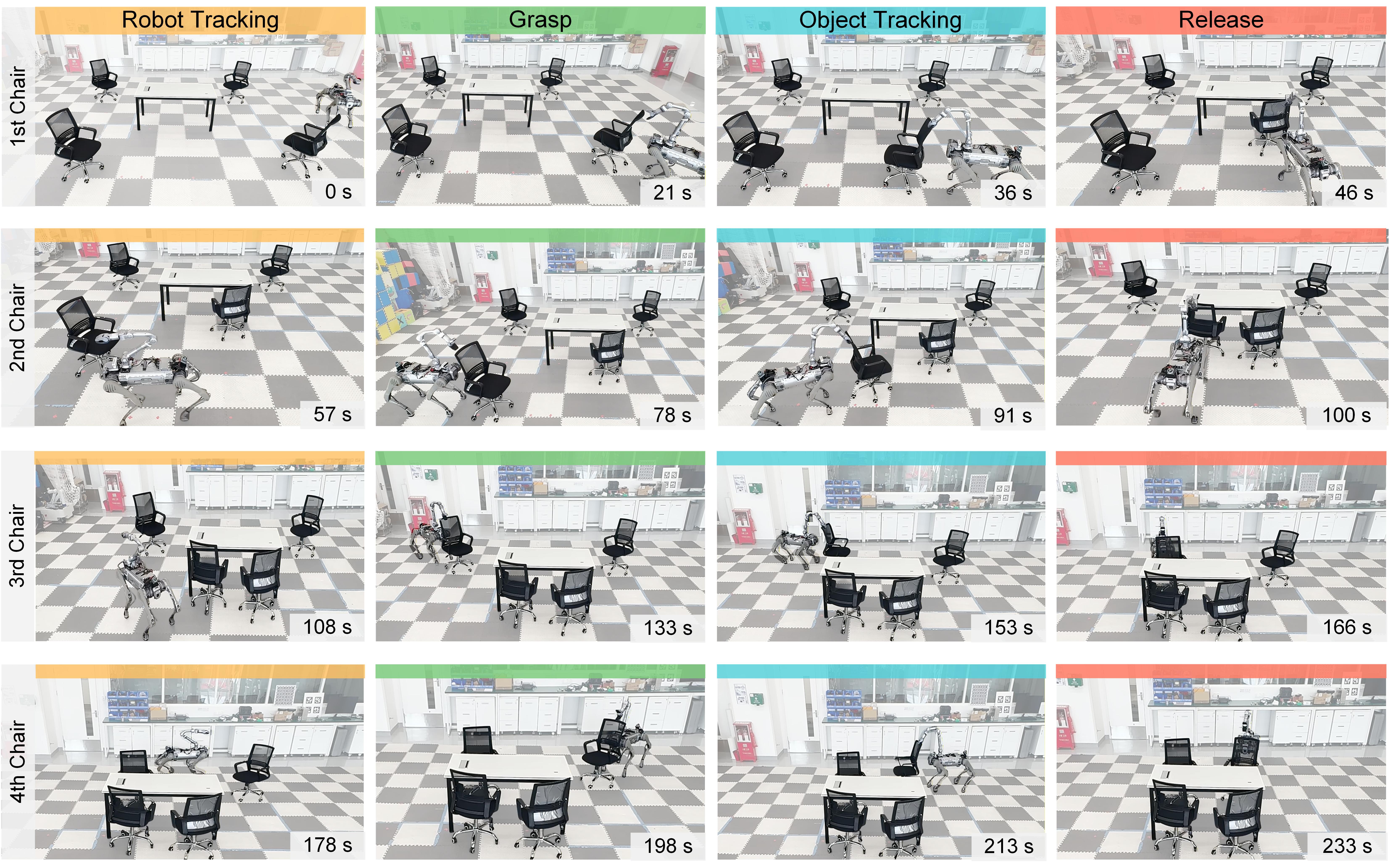}
	\setlength{\abovecaptionskip}{-0pt} 
	\caption{Illustration of continuous chair arrangement. Each row demonstrates four different stages of rearranging a single chair, including robot tracking, object grasping, object tracking, and releasing the object.}
	\label{fig:four_chair}
\end{figure*}

\subsubsection{Continuous Chair Rearrangement}
In this case, we randomly place four chairs with arbitrary positions and orientations in a 6\,m$\times$8\,m area and specify four target poses. The objective is to rearrange all four chairs to their corresponding target poses. For each chair, rearrangement is considered complete if the position error is below $0.3$\,m and the orientation error is below $45^\circ$. One example of the experimental results is shown in Fig. \ref{fig:four_chair}, where the target poses are located on both sides of a white office desk. Each chair rearrangement consists of four phases: tracking the robot trajectory, grasping the object, tracking the object trajectory, and releasing the object. The rearrangement of each chair takes less than 1 minute, and all four chairs are successfully rearranged in $233~\text{s}$. 
Furthermore, we demonstrate that our system can \textbf{continuously rearrange 32 chairs without a single failure} \textcolor{red}{(Supplementary Video S5)}, highlighting the strong robustness and reliability of the proposed system.

\begin{figure*}[t]	
	\centering
        % \vspace{3pt}
	\includegraphics[width=1.0\linewidth]{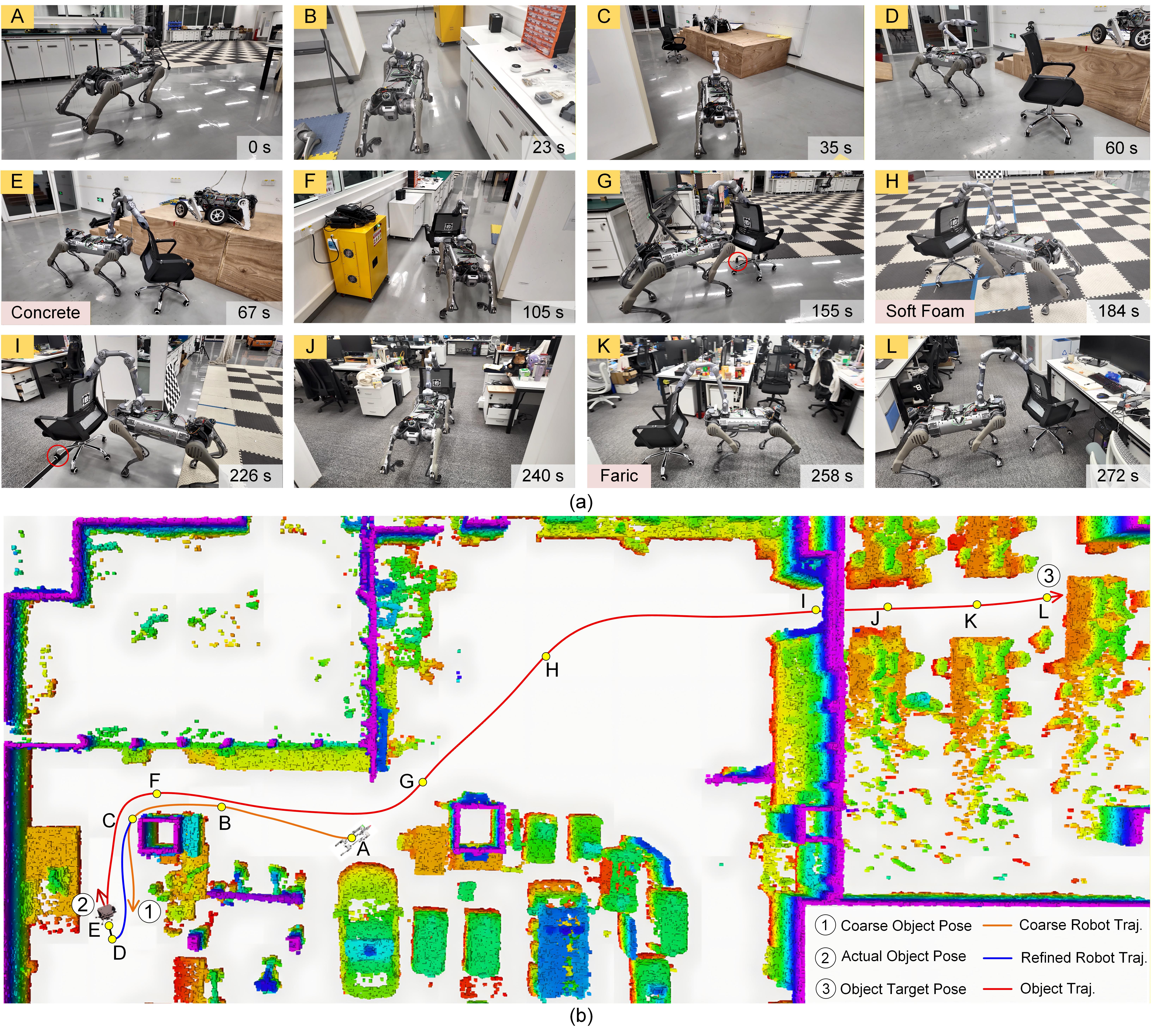}
	\setlength{\abovecaptionskip}{-18pt} 
	\caption{The legged manipulator performs long-distance autonomous chair rearrangement in a large indoor environment. (a) Some key frames during the arrangement. (b) The point cloud map and the generated trajectories, with the corresponding key waypoints in (a) highlighted in yellow.}
	\label{fig:auto_1}
\end{figure*}

\subsubsection{Autonomous Long-Distance Object Rearrangement}
As shown in Fig.~\ref{fig:auto_1}, the robot is given an approximate initial chair position \textcircled{\raisebox{-0.15ex}{1}} and a target pose \textcircled{\raisebox{-0.15ex}{3}}. Starting from an arbitrary location at \(t = 0\,\text{s}\) (A), it plans a coarse robot trajectory (orange) toward the approximate chair location. When the robot enters a 5\,m radius region centered at the initial chair position (B), the object detection module is activated. Once the chair is detected at \(t \approx 35\,\text{s}\) (C), the robot stops and estimates the chair's position and orientation relative to its base frame, then computes a pre-grasp pose and replans a fine robot trajectory (blue) to this pose, reaching it at \(t \approx 60\,\text{s}\) (D).
At the pre-grasp pose, the robot further refines the actual object pose \textcircled{\raisebox{-0.15ex}{2}} using the attached AprilTag and grasps the chair at \(t \approx 67\,\text{s}\) (E). It then plans a trajectory (red) for the coupled robot-chair system toward the target pose, and tracking with the proposed high-level object controller to accomplish the task by \(t \approx 272\,\text{s}\) (L). Throughout the process, the relative pose of the chair with respect to the robot is updated online through the real-time detection of the AprilTag pose on the chair using the camera. It is worth noting that this task is highly challenging. \textbf{The robot is required to move the chair along an approximately 40\,m route, traversing narrow passages without any collisions (F), traversing uneven terrain with protrusions (G and I), and moving across floor surfaces with varying materials, including concrete (F), soft foam (H), and fabric-covered ground (K).} These results further confirm that the proposed large-object rearrangement system enables robust, fully autonomous long-distance rearrangement. A video demonstration of this experiment is provided in \textcolor{red}{Supplementary Video S6}.

\section{Limitations}\label{sec:limit}
In this section, we discuss the limitations of the current approach and highlight directions for future research. First, in our current system, the approximate initial positions of objects and their target rearrangement poses are still assumed to be given a priori. We require commonsense knowledge to automatically determine target poses based on the environment. To this end, an interesting direction is to combine this work with large foundation models \cite{black2025pi_} that possess rich world knowledge, enabling the robot to autonomously explore the environment, understand scene semantics, and generate high-level goal poses and interaction strategies even in the absence of precise prior information.

Second, similar to \cite{li2025robotmover}, the graspable regions of the object are assumed to be known in advance. Although this simplification does not affect the main experimental results, an important avenue for future work is to integrate grasp affordance prediction into the RL-based controller \cite{ardon2019learning}, allowing the policy to discover feasible grasp or contact locations on previously unseen objects and to adapt its interaction strategy accordingly. 

Third, compared to prior object-specific policies trained for a single object, our approach makes a step forward by generalizing to previously unseen objects using a single policy. However, this generalization currently applies primarily to objects that induce similar robot-object interaction configurations. Performance may further degrade when encountering object-ground frictional models not covered during training. For instance, a policy trained on objects dominated by sliding friction may fail to transfer well to objects whose motion is governed by rolling friction. Actually, achieving truly object-agnostic generalization remains a long-standing and central challenge in robotics \cite{brunke2022safe}.
One potential way to extend generalization within the current framework is to train on objects exhibiting more diverse interaction configurations.
While the present work only considers three object categories, the framework is not inherently restricted to this setting and can be readily extended to a wider range of objects. But it requires accurate 3D object models, which are difficult to obtain and scale to the full diversity of real-world objects.
Inspired by recent advances in large language models \cite{brown2020language}, another promising direction is to leverage web-scale interaction data to learn general human-object interaction patterns and transfer them across embodiments to robotic systems \cite{zitkovich2023rt}.

\section{CONCLUSIONS}\label{sec:conclusion}
This study successfully demonstrates the first complete, long-horizon, and robust large-object rearrangement system with a legged manipulator that tightly integrates perception, planning, and whole-body object control. Building on a hierarchical design, a high-level RL-based object velocity controller is trained on top of a low-level WBC to enable versatile object manipulation. In addition, to accurately control different objects with only a single policy, we introduce two key modules. The ICR module abstracts the robot-object interaction configurations into a graph structure and provides a unified representation for the policy to better adapt to different objects. Besides, the object velocity estimator exploits historical proprioceptive data from the legged manipulator to predict object velocities, thereby improving tracking accuracy in a closed-loop manner without relying on external sensors. Extensive simulations and real-world experiments, including long-horizon tasks with various objects and large-scale environments, highlight that the proposed framework achieves robust, efficient, and safe object rearrangement performance. Overall, this work opens up new application scenarios for the legged manipulator and provides a flexible and effective paradigm for large-object rearrangement tasks. 
In the future, we plan to extend this framework to more versatile loco-manipulation capabilities and more sophisticated multi-object interaction scenarios, further advancing the ability of robots to serve humans in a wider range of real-world applications.

                                  % the text height of the previous page by a suitable amount.
                                  %, so it should come on the page before the last. Make
                                  % sure that you do not shorten the text height too much.

%%%%%%%%%%%%%%%%%%%%%%%%%%%%%%%%%%%%%%%%%%%%%%%%%%%%%%%%%%%%%%%%%%%%%%%%%%%%%%%%

%%%%%%%%%%%%%%%%%%%%%%%%%%%%%%%%%%%%%%%%%%%%%%%%%%%%%%%%%%%%%%%%%%%%%%%%%%%%%%%%

% \end{thebibliography}
% \small
\bibliographystyle{IEEEtran}
\bibliography{ref}

\end{document}